\documentclass[10pt,twocolumn,letterpaper]{fairmeta}

\newif\ifcvpr
\cvprfalse
\newif\ificml
\icmlfalse

\usepackage{xspace}
\RequirePackage{amsmath}
\RequirePackage{amssymb}
\makeatletter
\DeclareRobustCommand\onedot{\futurelet\@let@token\@onedot}
\def\@onedot{\ifx\@let@token.\else.\null\fi\xspace}

\def\eg{\emph{e.g}\onedot} 
\def\ie{\emph{i.e}\onedot} 
 
 \def\vs{\emph{vs}\onedot}
\def\wrt{w.r.t\onedot}

\makeatother

\ifcvpr  
\usepackage[dvipsnames]{xcolor}
\fi

\def\mypar#1{\vspace{0mm}\noindent\textbf{#1.}\hspace{0mm}}
\usepackage{array}
\newcolumntype{H}{>{\setbox0=\hbox\bgroup}c<{\egroup}@{}}

\usepackage{rotating} %
\usepackage{multirow}

\usepackage{bm}

\def\vc{{\bm{c}}}

\def\vr{{\bm{r}}}
\def\vx{{\bm{x}}}

\def\vi{{\bm{i}}}
\def\vg{{\bm{g}}}

\def\vG{{\bm{G}}}

\def\vC{{\bm{C}}}
\newcommand\norm[1]{\left\lVert#1\right\rVert}

\def\nlist{{K_\mathrm{IVF}}}
\def\nprobe{{P_\mathrm{IVF}}}
\newcommand{\nshort}{n_\mathrm{short}}

\makeatletter %
\DeclareRobustCommand\onedot{\futurelet\@let@token\@onedot}
\def\@onedot{\ifx\@let@token.\else.\null\fi\xspace}
\makeatother %

\def\eg{\emph{e.g}\onedot} 

\def\ie{\emph{i.e}\onedot}

\def\vs{\emph{vs}\onedot}
\def\wrt{w.r.t\onedot}

\definecolor{cvprblue}{rgb}{0.21,0.49,0.74}

\usepackage{sidecap}

\crefname{section}{Sec.}{Secs.}
\Crefname{section}{Section}{Sections}
\crefname{table}{Tab.}{Tabs.}
\Crefname{table}{Table}{Tables}
\crefname{figure}{Fig.}{Figs.}
\Crefname{figure}{Figure}{Figures}
\crefname{appendix}{App.}{App.}
\Crefname{appendix}{Appendix}{App.}

\newcommand{\OUR}{\textsc{QINCo}\xspace}
\newcommand{\IVFOUR}{\textsc{IVF-QINCo}\xspace}

\title{Residual Quantization\\
with Implicit Neural Codebooks}

\author[1,2,*]{Iris A. M. Huijben}
\author[1]{Matthijs Douze}
\author[1]{Matthew Muckley}
\author[2]{Ruud J. G. van Sloun}
\author[1]{Jakob Verbeek}

\affiliation[1]{FAIR at Meta}
\affiliation[2]{Eindhoven University of Technology}

\contribution[*]{Work done when interning at Meta.}

\abstract{Vector quantization is a fundamental operation for data compression and vector search. 
To obtain high accuracy, multi-codebook methods  represent each vector using codewords across several codebooks. 
Residual quantization (RQ) is one such method, which iteratively quantizes the error of the previous step. 
While the error distribution is dependent on previously-selected codewords, this dependency is not accounted for in conventional RQ as it uses a fixed codebook per quantization step.
In this paper, we propose QINCo, a neural RQ variant that constructs specialized codebooks per step that depend on the approximation of the vector from previous steps. 
Experiments show that QINCo outperforms state-of-the-art methods by a large margin on several datasets and code sizes. 
For example, QINCo achieves better nearest-neighbor search accuracy using 12-byte codes than the state-of-the-art UNQ using 16 bytes on the BigANN1M and Deep1M datasets.}

\date{\today}
\correspondence{Matthijs Douze (\email{matthijs@meta.com}) and Jakob Verbeek  (\email{jjverbeek@meta.com})}
\metadata[Code]{\url{https://github.com/facebookresearch/Qinco}}
\metadata[Note]{To appear at ICML 2024}

\begin{document}

\maketitle

\section{Introduction}

\begin{figure}
    \includegraphics[width=\linewidth,trim={0cm 6cm 15cm 0cm},clip]{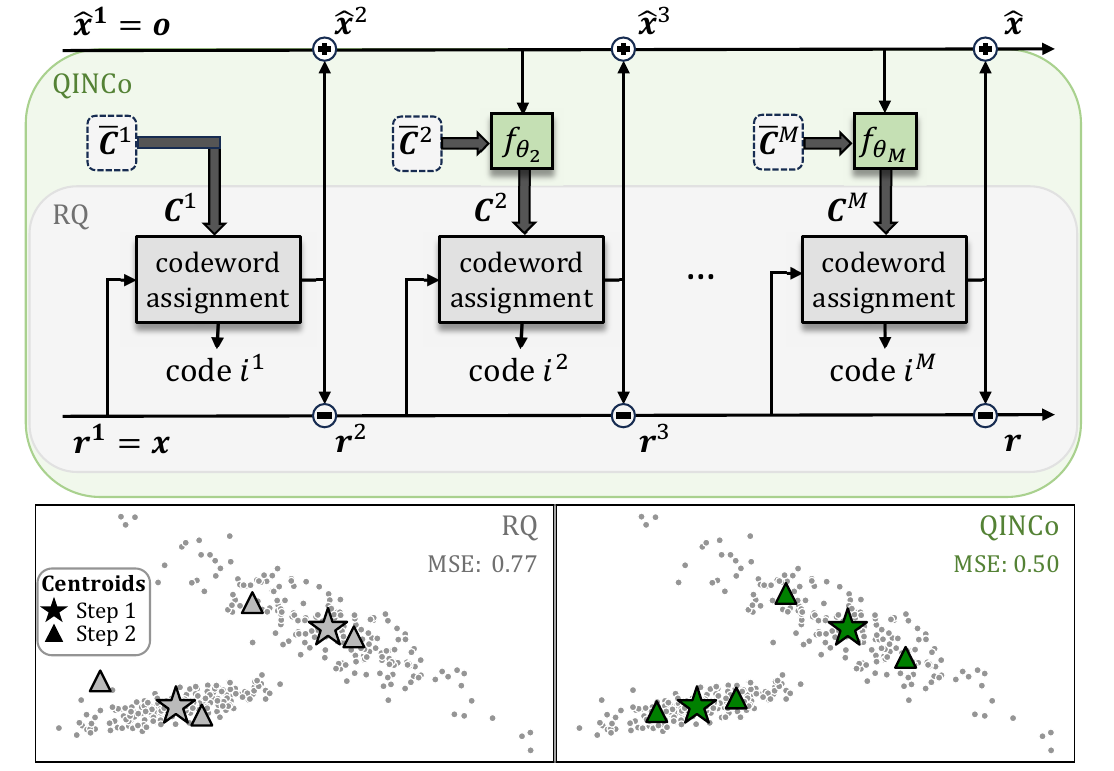}
    \vspace{-8mm}
    \caption{Top: Given a vector $\vx$,  RQ iteratively quantizes the residuals of previous quantization steps, using a single codebook $\vC^m$ for each step $m=1,\dots,M$.     
    \OUR extends  RQ  by using data-dependent codebooks that are implicitly parameterized via  a neural network $f_{\theta_m}$ that takes as input a base-codebook $\bar{\vC}^m$ and partial reconstruction $\hat{\vx}^m$ of the data vector $\vx$. 
    Bottom: Toy data example with $M\!=\!2$ quantization steps, each with $K\!=\!2$ centroids. In RQ, codebook centroids in the $2^{\text{nd}}$ level are independent of the $1^{\text{st}}$ level centroids, while \OUR adapts $2^{\text{nd}}$ level centroids to the residuals, reducing the mean-squared-error (MSE) by 35\%.
    } 
    \vspace{-2mm}
    \label{fig:architecture}
\end{figure}

Vector embedding is a core component of many machine learning systems for tasks such as analysis, recognition, search, matching, and others.
Part of the utility of vector embeddings is their adaptivity to different data modalities, such as text~\citep{schwenk2017learning,devlin2018bert,izacard2021contriever} and images~\citep{radford21clip,pizzi2022self,ypsilantis2023towards}.
In similarity search and recommender systems~\citep{paterek2007improving}, representing entities as vectors is efficient as it enables simple vector comparison. Many techniques and libraries have, nowadays, been developed to search through large collections of embedding vectors~\citep{malkov2018hnsw,guo2020accelerating,Morozov2019UnsupervisedSearch,douze2024faiss}. 

Vector embeddings can be extracted in different ways, \eg by taking the feature representation of a deep learning model.
After extraction, embeddings are typically compressed into fixed-length codes to improve efficiency for storage, transmission, and search. However, a fundamental trade-off exists, where shorter codes introduces higher distortion~\citep{cover91eit}, measured as the difference between the initial vector and its decoded approximation. In our work we focus on this vector compression process, and consider the data embedding approach itself as fixed.

A widespread technique to  compress embeddings is vector quantization (VQ)~\citep{gray1984vector}, which consists of representing each vector with a nearby ``prototype'' vector. 
Effective quantizers adapt to the data distribution by learning a codebook of centroids from a representative set of training vectors. 
The number of distinct centroids grows exponentially with the code size. 
The k-means VQ algorithm represents all centroids of the codebook explicitly. 
It tends to be near-optimal, but it does not scale to codes larger than a few bytes because of this exponential growth.
Multi-codebook quantization (MCQ) represents centroids as combinations of several codebook entries to avoid the exponential growth. 
Seminal MCQ techniques ---such as product quantization (PQ), residual quantization (RQ), and additive quantization (AQ)--- are based on clustering and linear algebra techniques~\citep{jegou2010product, chen2010approximate,babenko2014additive,martinez2016revisiting,martinez2018lsq++}, while more recent approaches rely on deep neural networks~\citep{yu2018product,liu2018deep,Morozov2019UnsupervisedSearch,wang2022contrastive,Niu2023ResidualSearch}.

Conventional RQ~\citep{chen2010approximate}, being a special case of AQ, iteratively quantizes the residual between the original vector and its reconstruction from the previous quantization steps.
Standard RQ methods use a fixed codebook for each quantization step.
This is sub-optimal, as the data distribution for the residuals is dependent on previous steps.
To address this, we propose a neural variant of RQ.
Our method adapts the codebooks at each quantization step using a neural network, leading to large reductions in error rates for the final compressed vectors.
We call our method \OUR for {\bf Q}uantization with {\bf I}mplicit {\bf N}eural {\bf Co}debooks. \Cref{fig:architecture} shows the conceptual difference between RQ and \OUR.

In contrast to earlier neural MCQ methods \citep{Morozov2019UnsupervisedSearch,Zhu2023RevisitingQuantization}, \OUR transforms the codebook vectors, rather than the vectors to be quantized. The similarity of \OUR to a standard RQ enables combining it with inverted file indexes (IVF)~\citep{jegou2010product} and re-ranking techniques for fast approximate decoding, making \OUR, as well, suitable for highly accurate large-scale similarity search. 
Our contributions are as follows: 
\begin{itemize}
  \setlength\itemsep{0em}
\item 
    We introduce \OUR, a neural residual vector quantizer that --- instead of using fixed codebooks --- adapts quantization vectors to the  distribution of residuals. It is stable to train and has few hyperparameters.
\item \OUR sets state-of-the-art performance for vector compression on multiple datasets and rates, and thanks to
    its compatibility with fast approximate search techniques, it also beats state-of-the-art similarity search performance for high recall operating points. 
\item 
    \OUR codes can be decoded from the most to the  least significant byte, with prefix  codes 
    yielding accuracy on par with 
    codes specifically trained for that code length, making \OUR an effective multi-rate codec. 
\end{itemize}

Code can  be found at \url{https://github.com/facebookresearch/QINCo}.

\section{Related Work}
\label{sec:related}

\mypar{Vector quantization}
A vector quantizer maps a vector $\vx\in \mathbb{R}^D$ to a vector taken from a finite set of size $K$~\citep{gray1984vector}. %
This set is called the codebook, and each codebook entry is referred to as ``centroid'',  or ``codeword''.
The objective is to minimize the distortion between $\vx$ and its quantization. %
Lloyd's algorithm, a.k.a.\ k-means, produces a set of codewords, leading to  codes of size $\lceil \log_2(K)\rceil$ bits. 
K-means, however, only scales well up to a few million centroids, resulting in code lengths in the order of 20 bits, which is too coarse for many applications.

\mypar{Multi-codebook quantization}
To scale beyond the inherent limitations of k-means, MCQ uses \emph{several} k-means quantizers, for which various approaches exist. PQ slices vectors into sub-vectors that are quantized independently~\citep{jegou2010product}. 
AQ, on the other hand, represents each vector as a sum of multiple codebook entries~\citep{babenko2014additive,martinez2016revisiting,martinez2018lsq++}, and RQ progressively quantizes residuals~\citep{chen2010approximate}. 
We build upon RQ, using neural networks to improve its accuracy by adapting codebooks to residual distributions.

\mypar{Neural quantization}
Neural quantization has been explored to learn discrete data representations, which can be used in discrete sequence models for the generation of images~\citep{VandenOordDeepMind,esser2021taming,Lee_2022_CVPR,chang22cvpr} and audio~\citep{copet2023simple}. Instead, in this work we focus on discrete representation learning for the purpose of compression and retrieval. Previous works have combined existing MCQ approaches, \eg PQ, with neural encoders for improving compression and/or retrieval of specific data modalities, like images~\citep{Agustsson2017,liu2018deep, yu2018product,klein2019end,jang2021self,wang2022contrastive,El-Nouby2022}, audio~\citep{defossez2023high,kumar2023high} and graph networks~\citep{he2023semisupervised}. Improvements in these works typically arise from adjustments in the learning objective or improving the optimization of MCQ using regularizers or relaxations, while not fundamentally changing the MCQ procedure itself. On the contrary, in this work, we focus on a fundamental new approach for MCQ, while assuming data embeddings are readily available and fixed.

Most similar to our work are 
UNQ~\citep{Morozov2019UnsupervisedSearch} and DeepQ~\citep{Zhu2023RevisitingQuantization}, who also focus on improving MCQ for already-embedded vectors, using neural networks.
Both models include a trainable data transformation that precedes the non-differentiable quantization step and, therefore, model the selected quantization vector as a sample from a categorical distribution, for which gradient estimators exist.
DeepQ leverages the REINFORCE estimator~\citep{glynn1990likelihood, williams1992simple} with additional control variates to reduce its variance, and UNQ uses the straight-through-Gumbel-Softmax estimator~\citep{Jang2017,Maddison2017} with carefully initialized and trainable Boltzmann temperatures~\citep{huijben2022review}. 
Both models use the nearest centroids, rather than a sampled centroid, for encoding after training. 
Opposed to UNQ and DeepQ, \OUR transforms the codebooks, rather than the data to be quantized, and thus encodes in the data space directly without leveraging a trainable transformation before quantization. This omits the need for gradient estimation. Moreover, it prevents posterior collapse after which all transformed embeddings are projected on the same centroid, something that requires additional regularization in training of UNQ and DeepQ.

\mypar{Re-ranking}
It is common practice to accelerate large-scale nearest neighbor search with approximation techniques that rely on a cheap distance measure to select a ``shortlist'' of nearest neighbors, which are subsequently re-ordered using a more accurate measure. This re-ordering can, \eg, be done using a finer quantizer~\citep{jegou2011searching} ---or in the limit without quantizer~\citep{guo2020accelerating}--- compared to the one used for creating the shortlist. It is also possible to \emph{re-interpret} the same codes with a more complex decoding procedure. 
For example, polysemous codes~\citep{douze2016polysemous} can be compared both as binary codes with Hamming distances, similar to~\citep{he2013kmeanshashing}, and as PQ codes. 
UNQ~\citep{Morozov2019UnsupervisedSearch} uses a fast AQ for search and re-ranks with a slower decoding network. 
It has also been shown that in some cases, given a codec, it is possible to train a neural decoder that improves the accuracy~\citep{Amara2022NearestPerspective}, and use the trained decoder to re-rank the shortlist. 
To enable fast search with \OUR, we also create a shortlist for re-ranking with a less accurate but faster linear decoder for which ---given the \OUR encoder--- a closed-form solution is available in the least-squared sense~\citep{babenko2014additive}.
On top of the approximate decoding, an inverted file structure (IVF) can direct the search on a small subset of database vectors. 
UNQ was extended in this way by~\citet{noh2023disentangled}. We show that the IVF structure integrates naturally with \OUR.

\mypar{Other connections} In our work a network is used to dynamically parameterize residual quantization codebooks. This is related to weight generating networks, see 
\eg \cite{ma2020weightnet}, and in a more remote manner to approaches that use one network to perform gradient-based updates of another network, see \eg \cite{andrychowicz16nips}.
\section{RQ with Implicit Neural Codebooks}
\label{sec:method}

We first briefly review RQ to set some notation;
for more details see, \eg,~\citet{chen2010approximate}.
We use $\vx\in \mathbb{R}^D$ to denote vectors we aim to quantize using $M$ codebooks of $K$ elements each.
Let $\hat{\vx}^m$ for $m=1,\dots, M$ be the reconstruction of $\vx$ based on the first $m-1$ quantization steps, with $\hat{\vx}^1:=\mathbf{0}$.
For each step $m$, RQ learns a codebook $\vC^m\in\mathbb{R}^{D\times K}$ to quantize the residuals $\vr^m=\vx-\hat{\vx}^m$.
We denote the centroids in the columns of $\vC^m$ as $\vc^m_k$ for $k=1,\dots,K$.
To encode $\vx$, at each step the selected centroid is $\vc^m_{i^m}$, where $i^m={\arg\min}_{k=1,\dots,K} \norm{\vr^m - \vc_k^m}^2_2$. The $M$ quantization 
indices $\vi=(i^1,\dots,i^M)$ are finally stored to represent $\vx$ using $M\lceil\log_2(K)\rceil$ bits. 
To decode $\vi$, the corresponding codebook elements are summed to obtain the approximation $\hat{\vx} = \sum_{m=1}^M \vc^m_{i^m}$.  

\subsection{Implicit neural codebooks}
At each step of the previously-described RQ scheme, all residuals are quantized with a single step-dependent codebook $\vC^m$. 
This is sub-optimal, as in general the distribution of residuals differs across quantization cells.
In theory, one could improve upon RQ by using a different specialized codebook for each hierarchical Voronoi cell.
In practice, however, as the number of hierarchical Voronoi cells grows exponentially with the number of quantization steps $M$, training and storing such local codebooks is feasible only for very shallow RQs.
For example, for short 4-byte codes with $M\!=\!4$ and $K\!=\!256$ we already obtain over four billion centroids.
Since training \emph{explicit} specialized codebooks is infeasible, we instead make these codebooks \emph{implicit}: they are generated by a neural network. The trainable parameters are not the codebooks themselves, but included in the neural network that generates them. 

For each quantization step $m$ we train a neural network $f_{\theta_m}$ that produces specialized codebook $\vC^m$ for the residuals $\vr^m$ in the corresponding hierarchical Voronoi cell. We condition $f_{\theta_m}$ upon the reconstruction so far $\hat{\vx}^m$, and a base 
codebook $\bar{\vC}^m$, and use it for
improving the $K$ vectors in the $m^{\text{th}}$ codebook in parallel:
$\vc_k^m=f_{\theta_m}(\hat{\vx}^m,\bar{\vc}_k^m)$. 
Base codebooks $\bar{\vC} = 
(\bar{\vC}^1,\dots,\bar{\vC}^M)$
are initialized using a pre-trained conventional RQ, and $f_{\theta_m}$ contains residual connections~\citep{he16cvpr} that let the base codebook pass-through, while allowing trainable multi-layer perceptrons (MLPs) to modulate the codebook. This architecture prevents spending many training cycles to achieve RQ baseline performance. 
The base codebooks are made trainable parameters themselves as well, so that $\bar{\vC}^m \subset \theta_m$.
See \cref{fig:architecture} for an  overview of \OUR and its relation to RQ.

More precisely, for all $K$ centroids in the $m^{\text{th}}$ codebook, $f_{\theta_m}$ first projects the concatenation of $\bar{\vc}_k^m$ and $\hat{\vx}^m$ using an affine transformation: $\mathbb{R}^{2D} \rightarrow \mathbb{R}^D$, after which $L$ residual blocks are used, each containing a residual connection that sums the input to the output of an MLP with two linear layers (ReLU-activated in between): $\mathbb{R}^D\rightarrow\mathbb{R}^h\rightarrow\mathbb{R}^D$. 
Since $\hat{\vx}^1=\mathbf{0}$ by construction, it does not provide useful context for conditioning, so we simply set $f_{\theta_1}$ to identity, resulting in $\vC^1 = \bar{\vC}^1$.
Therefore, the number of trainable parameters $|\theta| = \sum_{m}|\theta_m
|$ in \OUR equals: 
\begin{equation}
\label{eq:nr_parameters}
    |\theta| = M\underbrace{KD}_{\bar{\vC}^m} + (M-1)\big[\underbrace{\big(2D^2+D\big)}_{\text{concat. block}} + \underbrace{2LDh}_{\text{residual-MLPs}}\big].
\end{equation}

\subsection{Encoding, decoding and training}

Encoding a vector into a sequence of quantization indices proceeds as in conventional RQ encoding, with the only difference that \OUR constructs the $m^{\text{th}}$ codebook via $f_{\theta_m}$, instead of using a fixed codebook per step.  

As for decoding, unlike conventional RQ, \OUR follows a sequential process, as codebook-generating network $f_{\theta_m}$ requires partial reconstruction $\hat{\vx}^m$.
Given code $\vi$, for each quantization step $m=1,\dots,M$ reconstruction follows: $\hat{\vx}^{m+1} \leftarrow \hat{\vx}^m + f_{\theta_m}(\hat{\vx}^m,\bar{\vc}_{i^m}^m)$, with $\hat{\vx} := \hat{\vx}^{M+1}$ being the final reconstruction. 

To train parameters $\theta=(\theta_1,\dots,\theta_M)$ we perform stochastic gradient decent to minimize the mean-squared-error (MSE) between each residual and the selected centroid. 
For each quantization step, we optimize the following elementary training objective, defined per data point as:
\begin{equation}
    \mathcal{L}^m (\theta) = %
    \min_{k=1,\dots,K} \norm{\vr^m - f_{\theta_m}(\hat{\vx}^m,\bar{\vc}_k^m)}^2_2.
\end{equation}
\noindent Note that both $\vr^m$ and $\hat{\vx}^m$ implicitly depend on parameters $(\theta_1,\ldots,\theta_{m-1})$. Therefore, gradients from later quantization steps propagate back to earlier ones as well. 
Combining this loss for all $M$ steps yields the final loss: 
\vspace{-0.1cm}
\begin{equation}
\label{eq:loss}
    \mathcal{L}_\textrm{\OUR}(\theta) =  
   \sum_{m=1}^M \mathcal{L}^m (\theta).
\vspace{-0.1cm}
\end{equation}

\section{Large-scale Search with \OUR}
\label{sec:search}

For nearest-neighbor search in billion-scale datasets it is prohibitive to exhaustively decompress all vectors with \OUR, and compute distances between the query and the decompressed vectors. The resemblance of \OUR to conventional MCQ enables the use of existing methods to speed up similarity search. To this end, we introduce a fast search pipeline, referred to as \IVFOUR, that includes IVF (\cref{sec:ivfqinco}), approximate decoding (see \cref{sec:LSdecoder}), and re-ranking with the \OUR decoder. This pipeline gradually refines the search, and concentrates compute on the most promising database vectors. 

\subsection{Inverted file index (IVF)}
\label{sec:ivfqinco}

A common technique in large-scale search
consists of partitioning the database in $\nlist$ buckets using k-means,
and maintaining for each bucket a list of assigned vectors~\citep{jegou2010product}. 
Given a query, only data in the $\nprobe \ll \nlist$ buckets corresponding to the $\nprobe$ centroids closest to the query are accessed to speed up search.
In addition, since a database vector is assigned to a bucket, this means that the nearest centroid is the bucket centroid. 
This prior is used to make the codec more accurate~\citep{noh2023disentangled}.
IVF integrates naturally with \OUR: 
each database vector is assigned to one IVF bucket  $i^{\mathrm{IVF}}$, and that bucket's centroid is then used as the first estimate $\hat{\vx}^1=\vc_{i^{\mathrm{IVF}}}$ (instead of $\mathbf{0}$) of the \OUR code. 
Thus, in contrast to vanilla \OUR, the first codebook $\bar{\vC}^1$ is not fixed but generated by (non-identity) $f_{\theta_1}$. 
The subsequent \OUR coding steps remain the same.

\subsection{Approximate decoding}
\label{sec:LSdecoder}

Searching with IVF reduces the number of distance computations by a factor $\nlist / \nprobe$. 
However, compared to PQ and RQ, this does not result in competitive search times when combined with \OUR. 
This is because PQ and RQ, in addition to being cheaper to decode, can benefit from pre-computation of inner products between the query and all codebook elements. 
Distance computation between the query and a compressed database vector then reduces to summing $M$ pre-computed dot-products per database vector, which amounts to $M$ look-ups and additions~\citep{jegou2010product}.
Note that, for RQ, when using $\ell_2$ distances instead of dot-products for search, the norm of the vectors must also be stored~\citep{babenko2014additive}.

\OUR codebooks are not fixed, so this speed-up by table look-ups can not be applied directly.
However, it is possible to fit an additive decoder with fixed and explicit codebooks per quantization level, using codes from the \OUR encoder.
This returns approximate distances that can be used to create a short-list of database vectors for which the more accurate \OUR decoder is applied. 
More precisely, let $\vG=(\vG^1,\dots,\vG^M)$ denote a set of $M$ explicit codebooks, and let $\vg^m_k \in \mathbb{R}^D$ denote the $k^{\text{th}}$ element in the $m^{\text{th}}$ codebook. The MSE, defined per data point $\vx$, yields:
\begin{equation}
\mathcal{L}_\mathrm{MSE} (\vG) = ||\vx - \sum_{m=1}^M \vg^m_{i^m} ||_2^2  ,
\vspace{-0.05cm}
\end{equation}
where $\sum_{m=1}^M \vg^m_{i^m}$ is the reconstruction of $\vx$ using code $\vi$ from the \OUR encoder.
This optimization can be solved in closed form~\citep{babenko2014additive}.
We refer to this approximate decoder as ``AQ decoder''.

\subsection{Implementation}

We implement \IVFOUR in Faiss~\cite{douze2024faiss}, starting from a standard IVF index with AQ encoding. For each query, we use HNSW~\citep{malkov2018hnsw} to search the $\nprobe$ nearest centroids~\citep{baranchuk2018revisiting} and do compressed-domain distance computations in the corresponding inverted lists (note that, similar to RQ, this requires one additional byte per vector to encode the norms). 
We retrieve the top-$\nshort$ nearest vectors with approximate distances from the AQ decoder. Then we run \OUR decoding on the shortlist to compute the final results. 
See~\cref{app:ivfimplem} for more implementation details.

\section{Experiments}
\label{sec:experiments}

\subsection{Experimental setup}

\mypar{Datasets and metrics} 
We leverage datasets that vary in dimensionality ($D$) and modality: Deep1B~($D$=96)~\citep{babenko2016efficient} and BigANN~($D$=128)~\citep{jegou2011searching} 
are widely-used benchmark datasets for VQ and similarity search that contain CNN image embeddings and SIFT descriptors, respectively. %
Facebook SimSearchNet++ (FB-ssnpp; $D$=256)~\citep{simhadri2022results} contains image embeddings intended for image copy detection that were generated using the SSCD model~\citep{pizzi2022self} for a challenge on approximate nearest neighbor search. 
It is considered challenging for indexing, as the vectors are spread far apart. 
SIFT1M~($D$=128)~\citep{jegou2010product} is a smaller-scale dataset of SIFT descriptors used for vector search benchmarks. 
For all datasets, we use available data splits that include a database, a set of queries and a training set, and we hold out a set of 10k vectors from the original training set for validation, except for the smaller SIFT1M dataset for which we use 5k of the 100k vectors as validations vectors. Lastly, we introduce a new \emph{Contriever} dataset that consists of 21M 100-token text passages extracted from Wikipedia, embedded ($D$=768) using the Contriever model~\citep{izacard2021contriever}. This model is a BERT architecture~\citep{devlin2018bert} fine-tuned specifically for text retrieval. We randomly split the embeddings in 1M database vectors, 10k queries, and 20M training vectors, of which we use 10k as a hold-out validation set.

We report compression performance using MSE on 1M database vectors. 
To evaluate search performance we additionally report the nearest-neighbor recall percentages at ranks 1, 10 and 100 using 10k non-compressed queries and 1M or 1B compressed database vectors.
For resource consumption we focus on parameter counts:
since \OUR contains essentially linear layers, the decoding time is proportional to this count, making it a good proxy for run time.

\mypar{Baselines}
We compare \OUR to widely-adopted baselines
OPQ~\citep{ge2013optimized}, RQ~\citep{chen2010approximate}, and LSQ~\citep{martinez2018lsq++}, for which we use implementations in the Faiss library with default settings\citep{douze2024faiss}. 
We also compare to state-of-the-art neural baselines UNQ~\citep{Morozov2019UnsupervisedSearch}, RVPQ~\citep{Niu2023ResidualSearch}, and DeepQ~\citep{Zhu2023RevisitingQuantization}. 
RVPQ slices vectors into chunks like PQ and subsequently performs RQ separately in each block rather than using a single quantizer per block.
For UNQ, RVPQ and DeepQ we quote performance from the original papers. 
For UNQ we also reproduced results using the author's public  code, and run additional experiments, see \cref{app:experiments_with_UNQ} for more details.

\begin{table}
    \caption{Comparison of \OUR with state-of-the-art methods in terms of reconstruction error (MSE)  and nearest-neighbor search recall (R@1) in percentages. We report \OUR with $L\!=\!16$, except for Contriever1M, where $L\!=\!12$ is used.
}
    \centering
    \resizebox{\columnwidth}{!}{
    \setlength{\tabcolsep}{3.1pt}
  \begin{tabular}{clcccccccc}
  \toprule
     && 
     \multicolumn{2}{c}{\bf BigANN1M }&  \multicolumn{2}{c}{\bf Deep1M} &\multicolumn{2}{c}{\bf Contriever1M}& \multicolumn{2}{c}{\bf FB-ssnpp1M} \\
     \cmidrule(lr){3-4}\cmidrule(lr){5-6}\cmidrule(lr){7-8}\cmidrule(lr){9-10}
     && MSE & R@1&  MSE & R@1& MSE & R@1&  MSE & R@1 \\
     && ($\times 10^4$) &&&&&&($\times 10^4$)  \\
     \midrule
\multirow{5}{*}{\rotatebox{90}{\bf8 bytes}} 
&OPQ&                                       2.95 & 21.9 &  0.26 & 15.9 &  1.87 & 8.0    & 9.52 & 2.5 \\
&RQ&                                        2.49 & 27.9 &  0.20 & 21.4 &  1.82 & 10.2   & 9.20 & 2.7 \\
&LSQ&                                       1.91 & 31.9 &  0.17 & 24.6 &   1.65 & 13.1  & 8.87& 3.3  \\
&UNQ  & 1.51 & 34.6 &  0.16 & 26.7 &  --- &  --- &  --- &  ---  \\
&\OUR &                                     \bf 1.12 & \bf 45.2 & \bf 0.12 & \bf 36.3 & \bf1.40 & \bf 20.7 & \bf 8.67 & \bf 3.6 \\
     \midrule
     \multirow{5}{*}{\rotatebox{90}{\bf16 bytes}}
&OPQ&                                       1.79 & 40.5 & 0.14 & 34.9  & 1.71 & 18.3 & 7.25 & 5.0 \\
&RQ&                                        1.30 & 49.0 & 0.10 & 43.0  & 1.65 & 20.2 & 7.01 & 5.4 \\
&LSQ&                                       0.98 & 51.1 & 0.09 & 42.3  & 1.35 & 25.6 & 6.63 & 6.2 \\
&UNQ  & 0.57 & 59.3 & 0.07 & 47.9  &  --- &  --- & --- &  ---  \\
&\OUR &                                     \bf 0.32 &\bf 71.9 & \bf 0.05 & \bf 59.8 & {\bf 1.10}  & {\bf 31.1}  & \bf6.58 & \bf 6.4 \\
    \bottomrule
    \end{tabular}
     }
    \label{tab:best_db1Mresults}
    \vspace{-0.25cm}
\end{table}
\mypar{Training details}
We train models on 500k or 10M vectors (except for SIFT1M, that contains only 95k training vectors), and perform early stopping based on the validation loss.
During training, all data is normalized by dividing the vector components by their maximum absolute value in the training set. \Cref{app:training_details} provides additional training details.

The number of trainable parameters in \OUR scales linearly with the number of residual blocks $L$ and the hidden dimension $h$ of the residual-MLPs. 
Preliminary experiments showed that the performance gain of increasing either $L$ or $h$ by the same factor, was very similar, see \cref{app:L_vs_h}. Therefore, to vary the capacity of \OUR, we varied the number of residual blocks $L$, and fixed the hidden dimension to $h=256$.
For most experiments we use $M\in\{8,16\}$ quantization levels and vocabulary size $K=256$, which we denote as ``8 bytes'' and ``16 bytes'' encoding.

\begin{table}
    \caption{Recall values at different ranks for similarity search. \OUR with $L\!=\!4$ is reported.}
   \resizebox{\columnwidth}{!}{
    \centering
    \scriptsize
    \begin{tabular}{lccccccc}
    \toprule
    & \multicolumn{3}{c}{\bf 4 bytes} & \multicolumn{3}{c}{\bf 8 bytes} \\
    \cmidrule(lr){2-4}\cmidrule(lr){5-7}
    & R@1 & R@10 & R@100 & R@1 & R@10 & R@100 \\
    \midrule
  &\multicolumn{6}{c}{\bf SIFT1M} \\
  \midrule
    RVPQ  & 10.2 & 34.7 & 74.5 & 30.3 & 73.8 & 97.4 \\
    DeepQ  & 11.0 & 37.7 & 76.8 & 28.0 & 70.2 & 96.4 \\
    \OUR & \bf 14.9 & \bf 45.5 & \bf 82.7 & \bf 35.8 & \bf 80.4 & \bf 98.6\\
    \midrule
  &\multicolumn{6}{c}{\bf Deep1M} \\
  \midrule
    DeepQ  & 7.4 & 30.0 & 72.5 & 20.9 & 62.1 & 94.1 \\
    \OUR & \bf 9.1 & \bf 36.3 & \bf 77.8 & \bf 25.4 & \bf 72.1 & \bf 97.4 \\
    \bottomrule
    \end{tabular}
    }
    \label{tab:T100k_db1Mresults}
    \vspace{-0.25cm}
\end{table}

\subsection{Quantization performance}
\label{sec:results}

In \cref{tab:best_db1Mresults} we compare \OUR against the baselines on four datasets. 
For Contriever we report \OUR with $L\!=\!12$, for the other datasets we report $L\!=\!16$. 
\OUR outperforms all baselines on all datasets with  large margins.
On BigANN for example, \OUR reduces the MSE by 26\% and 44\% for 8 and 16 bytes encodings respectively, and search recall (R@1) is improved by more than 10 points for both encodings.
In general we find that \OUR optimally uses all codewords without explicitly enforcing this using regularization during training, see \cref{app:entropy}. Note that the methods that we compare have different numbers of parameters and training set sizes, and also vary in encoding and decoding speed. 
These factors are analyzed in \Cref{sec:search_exp,sec:additional_analyzes}. 

To compare to reported results for DeepQ~\citep{Zhu2023RevisitingQuantization} and RVPQ \citep{Niu2023ResidualSearch},
we train a smaller \OUR ($L\!=\!4$) on 100k vectors for Deep1B and 95k vectors for SIFT1M.
\Cref{tab:T100k_db1Mresults} shows that \OUR substantially outperforms these methods as well on both datasets.

\begin{figure}
    \centering
    \includegraphics[width=1.\columnwidth,trim={0cm 0.2cm 0cm 0.8cm},clip]{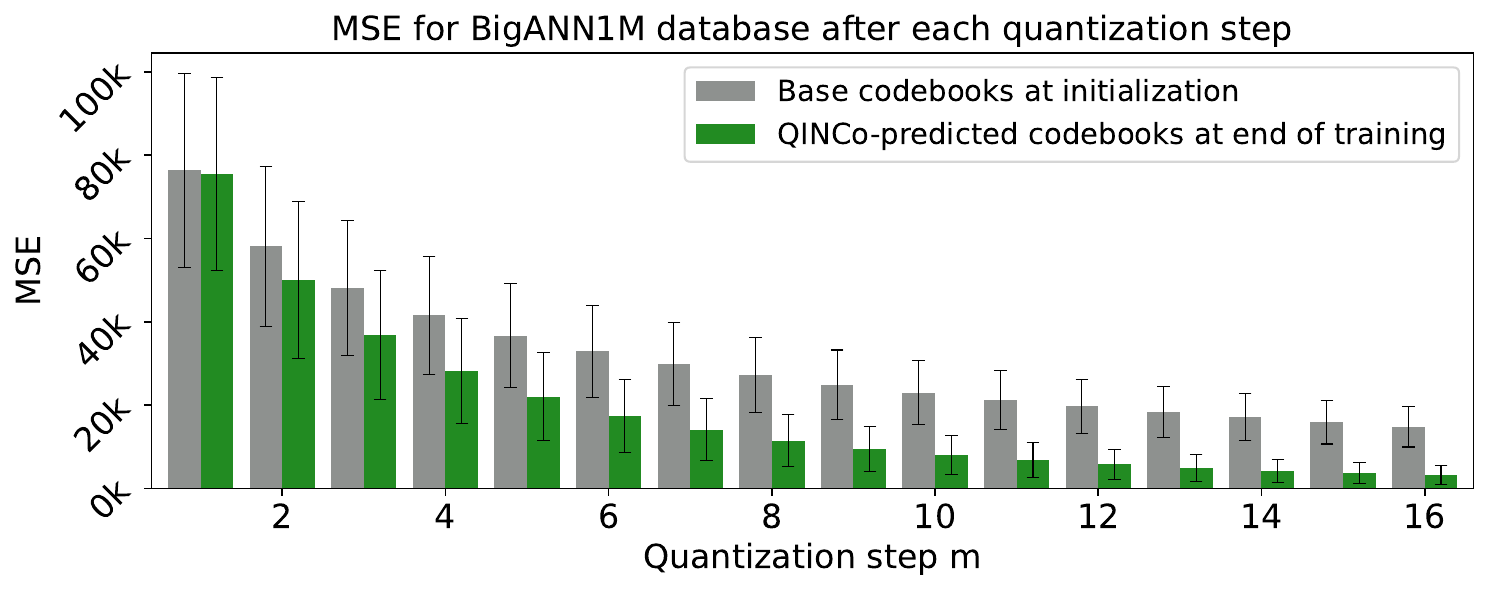}
    \caption{
MSE (mean $\pm$ std.~dev.) on BigANN1M across 16 quantization steps before training of \OUR ($L\!=\!16$), and after training on 10M samples.
     }
    \label{fig:MSE_perM}
    \vspace{-0.2cm}
\end{figure}

\Cref{fig:MSE_perM} shows that \OUR gains accuracy with respect to the base RQ in all quantization steps, but the relative improvement is larger in the deeper ones. 
An explanation is that for deeper quantization steps, the residual distributions tend to become more heterogeneous across cells, so specialized codebooks predicted by \OUR become more useful.

\subsection{Search performance} 
\label{sec:search_exp}

\begin{table}
    \caption{Complexity of encoding and decoding per vector (in floating point operations, FLOPS) and indicative timings on 32 CPU cores (in $\mu$s)
    with parameters: $D$=128; \OUR: $L$=2, $M$=8, $h$=256; UNQ: $h'$=1024; $b$=256; RQ: beam size $B$=5.     
    In practice, at search time for OPQ and RQ  we perform distance computations in the compressed domain, which takes $M$ FLOPS (0.16~ns).
}
       \resizebox{\columnwidth}{!}{
    \centering
    \scriptsize
    \setlength{\tabcolsep}{1.7mm}
    \begin{tabular}{lcrcr}
    \toprule
     & \multicolumn{2}{c}{\bf Encoding} & \multicolumn{2}{c}{\bf Decoding}  \\
    \cmidrule(lr){2-3}\cmidrule(lr){4-5}
    & FLOPS & time & FLOPS & time \\
    \midrule
       OPQ & $D^2 + KD$ &  1.5 & $D(D + 1)$   & 1.0 \\
       RQ  & $KMDB$   & 8.3 & $MD$  & 1.3 \\
       UNQ & $h'(D\!+\!h'\!+\!Mb\!+\!MK)$& 18.8  & $h'(b\!+\!h'\!+\!D\!+\!M)$  & 13.0  \\
    \OUR   & $2MKD(D+Lh)$ & 823.4 & $2MD(D+Lh)$& 8.3  \\
    \bottomrule
    \end{tabular}
    }
    \label{tab:complexity}
\end{table}

In \cref{tab:complexity} we report the complexity and corresponding encoding/decoding times of \OUR and baselines. 
All timings are performed on 32 threads of a 2.2~GHz E5-2698 CPU with appropriate batch sizes. 
In particular for encoding, \OUR is slower than the competing methods both in terms of complexity and  timings.
Given the encoding complexity of \OUR on CPU, we run encoding on GPU for all \OUR experiments not related to timing. 
The encoding time for the same \OUR model on a Tesla V100 GPU is 28.4~$\mu$s per vector.

Since the search speed depends on the decoding speed of the model, we experiment with approximate decoding for \OUR, as described in \cref{sec:LSdecoder}. 
For each query we fetch $\nshort$ results using the approximate AQ decoding and do a full \OUR decoding on these to produce the final search results. 
\Cref{tab:reranking} shows that the R@1 accuracy of the approximate AQ decoding is low compared to decoding with \OUR (and compared to RQ). 
However, re-ranking the top-1000 results (\ie, 0.1\% of the database) of the AQ decoder with \OUR brings the recall within 0.3\% of exhaustive \OUR decoding.

\begin{table}
    \caption{Search  accuracy (R@1) using the approximate AQ decoder only (row 1), AQ in combination with \OUR (with $L\!=\!2$) to re-rank a shortlist of size $\nshort$ obtained using the AQ decoder (rows 2, 3, 4), and \OUR to decode the full database (row 5).
    }
    \centering
    \scriptsize
    \begin{tabular}{lcccc}
    \toprule       
 &   \multicolumn{2}{c}{\bf BigANN1M} &     \multicolumn{2}{c}{\bf Deep1M} \\
     \cmidrule(lr){2-3}\cmidrule(lr){4-5}
  &  8 bytes & 16 bytes & 8 bytes & 16 bytes \\
    \midrule    
  AQ            & 12.7	& 15.6	& 11.9	& 17.6  \\
 $\nshort=10$  & 30.5	& 43.1	& 25.3	& 40.3  \\
$\nshort=100$ & 38.9	& 62.8	& 30.3	& 53.0  \\
 $\nshort=1000$& 40.1	& 67.2	& 31.2	& 54.9  \\
  \OUR             & 40.2	& 67.5	& 31.1	& 55.0  \\
\bottomrule
    \end{tabular}
    \label{tab:reranking}
    \vspace{-0.2cm}
\end{table}

Only using approximate decoding to create a shortlist does not yield competitive search speeds yet. 
As such, we experiment with \IVFOUR on billion-scale datasets, which combines AQ approximate decoding with IVF (see \cref{sec:search}).
We use \IVFOUR with $\nlist$=$10^6$ buckets. %
In terms of pure encoding (\ie without AQ decoding), \IVFOUR already improves the MSE of regular \OUR thanks to the large IVF quantizer, see \cref{tab:qinco_vs_ivfqinco}. 

\begin{table}[t]
\centering
\caption{
MSE of \OUR and \IVFOUR for 8- and 16-byte codes on BigANN1M for $L\!=\!4$.
}
\scriptsize 
\begin{tabular}{lcc}
\toprule
                & \bf 8 bytes & \bf 16 bytes \\
\midrule
QINCo           & $1.24\times 10^4$ & $3.77\times 10^3$ \\
IVF-QINCo        & $0.78\times 10^4$  & $2.74\times 10^3$ \\
\bottomrule
\end{tabular}
\label{tab:qinco_vs_ivfqinco}
\end{table}

\begin{figure}
    \hspace*{-4mm}
    \includegraphics[width=1.05\columnwidth,trim={0cm 0.2cm 0cm 0.1cm},clip]{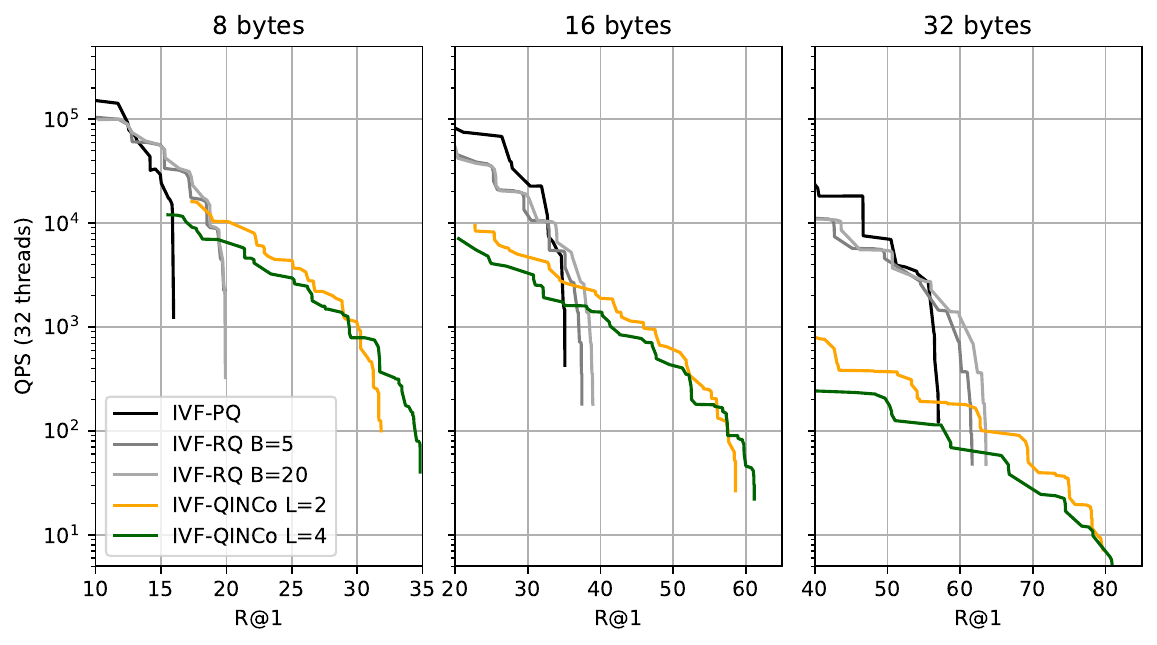}
    \caption{
    Speed-accuracy trade-off in terms of queries per second (QPS) and recall@1 for \IVFOUR, on BigANN1B ($10^9$ vectors), compared to IVF-PQ and IVF-RQ. 
    }
    \label{fig:IVFplot}
\end{figure}

\begin{figure*}
    \centering
    \includegraphics[width=1\linewidth]{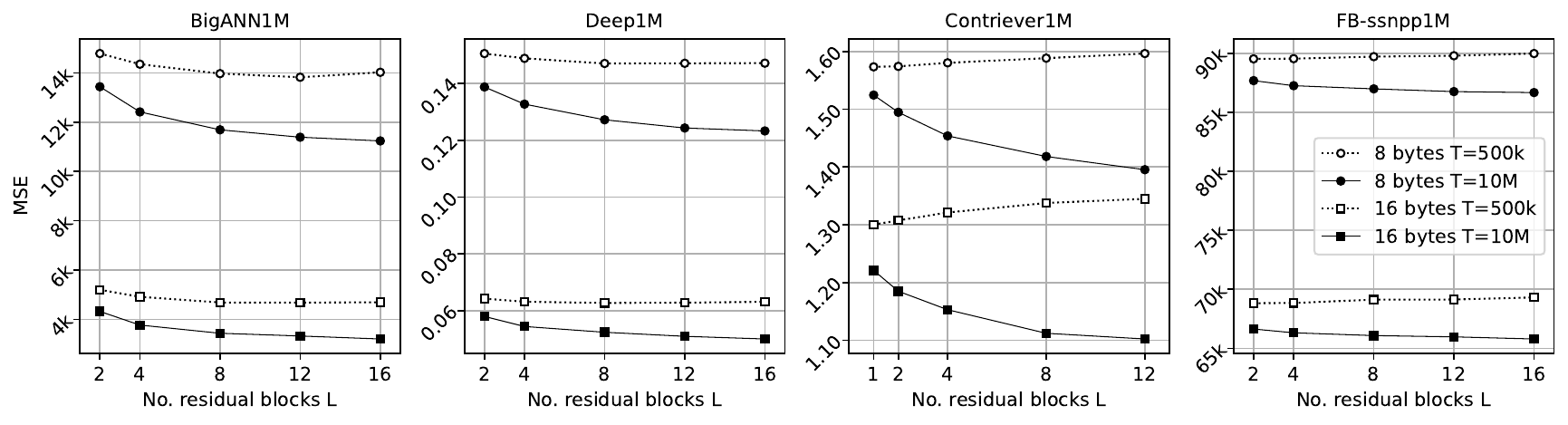}
    \caption{Performance of \OUR of  residual blocks $L$ and a training set size $T$ of 500k (open) or 10M (solid).
    }
    \label{fig:scaling_sweep}
    \vspace{-0.2cm}
\end{figure*}

In \cref{fig:IVFplot} we plot the speed-accuracy trade-offs obtained on BigANN1B (database of size $10^9$) using \IVFOUR, IVF-PQ and IVF-RQ. 
We report IVF-RQ results and IVF-QINCo with two settings of build-time parameters (number of blocks $L$ for IVF-QINCo and beam size $B$ for IVF-RQ) that adjust the trade-off between search time and accuracy.
There are three search-time parameters: $\nprobe$, $\mathrm{efSearch}$ (a HNSW parameter) and $\nshort$. 
For each method we evaluate the same combinations of these parameters and plot the Pareto-optimal set of configurations. 
We observe that there is a continuum from IVF-PQ, via IVF-RQ to \IVFOUR: IVF-PQ is fastest but its accuracy saturates quickly, IVF-RQ is a bit slower but gains about 5 percentage points of recall; 
\IVFOUR is again slower but results in 10 to 20 percentage points of recall above IVF-RQ. 
The impact of the build-time parameters is significant but does not bridge the gap between the methods. 
For the operating points where \IVFOUR is interesting, %
it can still sustain hundreds to thousands of queries per second. 
This is the order of speeds at which hybrid memory-flash methods operate~\citep{subramanya2019diskann}, except that \OUR uses way less memory. 
\Cref{app:fast_search} presents additional analyses on fast search with \IVFOUR.

\subsection{Further analyses}
\label{sec:additional_analyzes}

\mypar{Scaling experiments}
To investigate the interaction between training set size and model capacity, we train \OUR on both 500k and 10M vectors for codes of  8 and 16 bytes, and vary the number of residual blocks $L$. 
\Cref{fig:scaling_sweep} shows that in all cases the accuracy significantly improves with more training data, and that given enough training data it keeps improving with larger model capacity $L$. 
For less training data (500k vectors), increasing the capacity too much can degrade the accuracy, due to overfitting.

To test whether  baselines benefit similarly from more training data, we train OPQ, RQ and LSQ on 10M training vectors. \Cref{tab:scaling_training_vecs_FAISSbaselines} in \cref{app:scaling_unq} shows that these algorithms hardly benefit from more training data. 
UNQ was originally trained on 500k training vectors using shallow encoder and decoder designs: both only contained a two-layer MLP with $h'\!=\!1024$ hidden dimensions. 
By increasing either the depth ($L'$) or width ($h'$) of these MLPs, while training on 500k vectors, we found that UNQ suffered from overfitting and test performance decreased (also when deviating from the hyperparameter settings given by the authors). 
However, training UNQ on 10M vectors improved the MSE for deeper (larger $L'$) and wider (higher $h'$) MLPs. However, when evaluating the number of trainable parameters against MSE performance, \cref{fig:scaling_unq} in \cref{app:scaling_unq} shows that the Pareto front of these better-performing UNQ models remains far from \OUR's performance.

\begin{figure}
    \centering
    \includegraphics[width=\columnwidth]{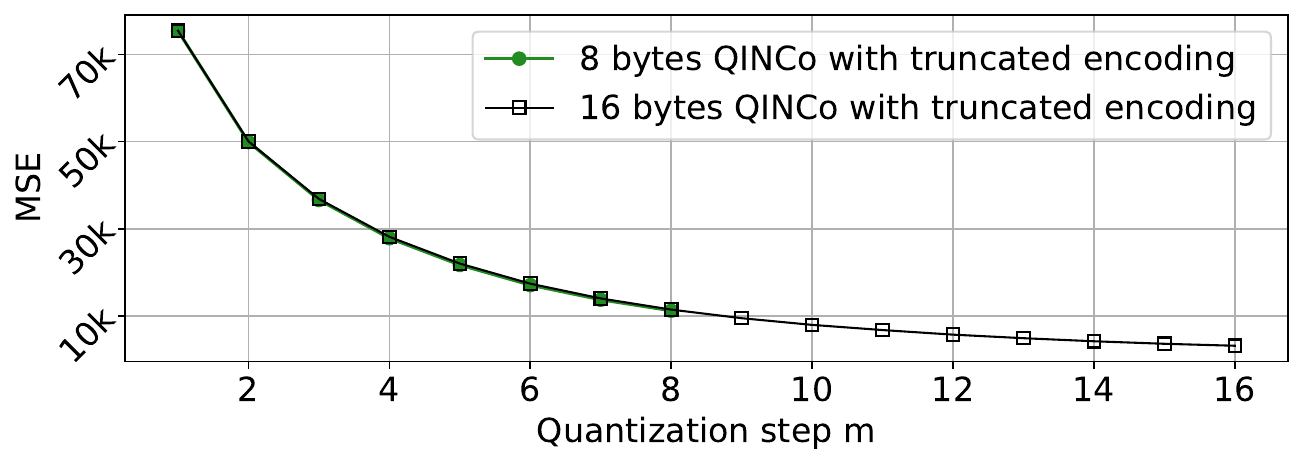}
    \caption{
     The MSE after the $m^\text{th}$ quantization step is very similar for the 8 bytes and 16 bytes models for BigANN1M.}
    \label{fig:cropping_codes}
    \vspace{-0.3cm}
\end{figure}

\mypar{Dynamic Rates}
We evaluate whether a \OUR model trained for long codes can be used to generate short codes, or equivalently, whether partial decoding can be performed by stopping the decoding after $m<M$ steps.
\Cref{fig:cropping_codes} shows the MSE per quantization step on BigANN1M for both the 8- and 16-byte models ($L\!=\!16$), which is almost identical for $m\leq8$.
This has several benefits: 
compressed domain rate adjustment (vectors can be approximated by cropping their codes);
amortized training cost by only training for the largest $M$;
and simple model management (only a single model is required). 
This also implies that the loss at step $m$ hardly influences the trainable parameters in steps $<m$. 
\Cref{app:cropping_codes} shows similar graphs for Deep1M and the R@1 metric for both datasets. 
They show that with 12 bytes and more, \OUR outperforms 16-byte-UNQ's R@1=59.3\% for BigANN1M and R@1=47.9\% for Deep1M. %

\begin{figure}
    \centering
    \includegraphics[width=0.8\columnwidth]{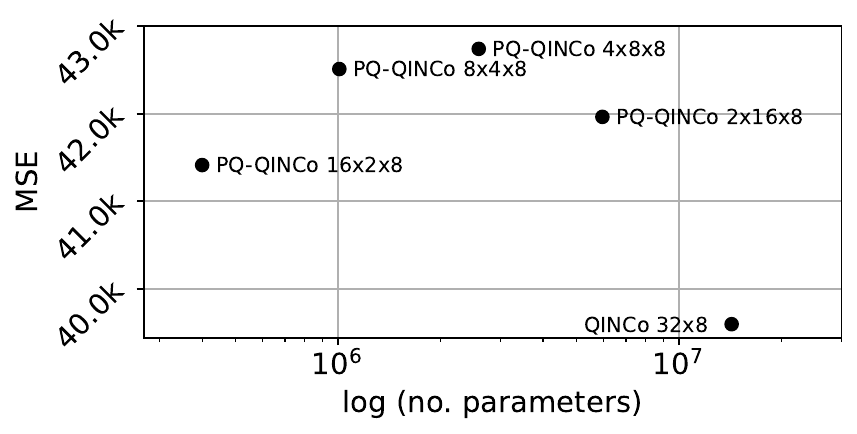}
    \vspace{-0.2cm}
    \caption{Comparing  32 byte   encodings of FB-ssnpp for \OUR and  PQ-\OUR.  
    The setting $16\times2\times8$ means we use 16 PQ blocks, $M\!=\!2$ residual steps and $K\!=\!2^8\!=\!256$ centroids.
    }
    \label{fig:PQQINCo_fbssnpp}
    \vspace{-0.5cm}
\end{figure}

\mypar{Integration with product quantization}
For efficiency when generating large codes, RQ is often combined with PQ to balance sequential RQ stages with parallel PQ coding~\citep{Babenko2015TreeClassification,Niu2023ResidualSearch}. 
In this setup, the vector is divided into sub-vectors, and an RQ is trained on each sub-vector. 
\OUR can equivalently be combined with PQ. 
We train \OUR and PQ-\OUR ($L\!=\!2$) on 10M vectors of FB-ssnpp for 32-byte encoding. 
\Cref{fig:PQQINCo_fbssnpp} shows the trade-off between number of parameters and performance for PQ-\OUR and \OUR. 
Interestingly, using more PQ blocks deteriorates performance until a turning point, where performance improves again. 
Vanilla PQ~\citep{jegou2010product} has  65.5k trainable parameters (way fewer than the PQ-\OUR variants) and obtains MSE=55.7k (much worse than PQ-\OUR).
Compared to \OUR, PQ-\OUR speeds up encoding and search in high-rate regimes, at the cost of accuracy.

\mypar{\bf \OUR variant for high-dimensional data}
The number of trainable parameters in \OUR scales in $\mathcal{O}(D^2)$, see \cref{eq:nr_parameters}. 
For high-dimensional embeddings, we propose \OUR-LR, a variant of \OUR that contains an additional low-rank (LR) projection: 
for each \OUR step, we replace the first affine layer $\mathbb{R}^{2D} \rightarrow \mathbb{R}^{D}$ by two linear layers that map $\mathbb{R}^{2D} \rightarrow \mathbb{R}^{h} \rightarrow \mathbb{R}^{D}$. 
\OUR-LR scales in $\mathcal{O}(hD)$.
We fix $h\!=\!256$ (same as the residual blocks) and observe that \OUR-LR (8 bytes; $L\!=\!4$) trained on 10M Contriever embeddings achieves a database MSE of 1.46 with 16.71M trainable parameters, as compared to an MSE of 1.45 for vanilla QINCo with 20.85M parameters. \OUR-LR is thus 20\% more parameter-efficient, while barely loosing performance, making
\OUR-LR interesting for even larger embeddings, as more than 1,000 dimensions is not uncommon~\citep{devlin2018bert,oquab2023dinov2}.

\mypar{Allocating bits}
Given a fixed bits budget $M\log_2(K)$, PQ and additive quantizers are more accurate with a few large codebooks (small $M$, large $K$) than with many small codebooks (large $M$, small $K$), as the latter setting has a lower capacity (fewer trainable parameters). 
To investigate whether \OUR behaves similarly, we trained \OUR ($T=500\textrm{k},L\!=\!4$, and a base learning rate of $10^{-3}$) on \mbox{BigANN1M} with $M\!=\!10$ codebooks with the default $K\!=\!2^8$; and $M\!=\!8$ codebooks with $K\!=\!2^{10}$. 
\Cref{tab:10x8_vs_8x10} shows that these two modes of operation are more similar, \ie only 2.1\% decrease in MSE, than for RQ and LSQ, for which MSE decreased 11.1\% and 6.5\%, respectively. 

The reason for this different behavior of \OUR with respect to additive quantizers, is that the relation between $M$, $K$ and the number of trainable parameters in \OUR depends on the number of residual blocks $L$. For increasing $L$, the two modes of operation (small $M$, large $K$ vs small $K$, large $M$) get closer in terms of trainable parameters, which reduces the gap in performance.

\begin{table}
  \caption{Performance trade-offs on BigANN1M  for two \OUR settings that yield 10-byte codes.} %
   \resizebox{\columnwidth}{!}{
\scriptsize
    \centering
    \setlength{\tabcolsep}{1.6mm}
    \begin{tabular}{lcccc|cc}
    \toprule
          &\multicolumn{2}{c}{$\mathbf{M\!=\!10}$, $\mathbf{K\!=\!2^{8}}$} &\multicolumn{2}{c}{$\mathbf{M\!=\!8}$, $\mathbf{K\!=\!2^{10}}$} \\  
         \cmidrule(lr){2-3}\cmidrule(lr){4-5}
         &  MSE ($\times\!10^4$) & R@1 & MSE ($\times\!10^4$) & R@1 & $\Delta$ MSE & $\Delta$ R@1\\
          \midrule
         RQ&  2.07 & 35.5 & 1.84 & 37.2 & -11.1\% & +4.8\%\\
         LSQ&  1.55 & 37.6 & 1.45 & 39.3 & -6.5\% & +4.5\%\\
         \OUR &  0.96 & 49.9 & 0.94 & 50.1 & -2.1\% & +0.4\% \\
          \bottomrule
    \end{tabular}
    }
    \label{tab:10x8_vs_8x10}
    \vspace{-0.3cm}
\end{table}

\mypar{Additional ablations studies}
Finally, we summarize main findings from more ablations presented in~\cref{app:ablations}.

(i) \OUR can be trained using only the MSE loss after the last quantization step, \ie $\mathcal{L}^M(\theta)$, instead of summing the $M$ losses from all quantization steps as in \cref{eq:loss}. 
However, this drastically reduced performance and the optimization became unstable.

(ii) \OUR's $M$ losses can be detached, such that each loss only updates the parameters $\theta_m$ of one \OUR step. 
This slightly deteriorated or did not affect MSE, while recall levels remained similar, or slightly improved in some cases. 
In general, each loss thus has a marginal impact on earlier quantization steps. 
This corroborates our finding that \OUR can be used with dynamic rates during evaluation.

(iii) The number of trainable parameters in \OUR scales linearly with the number of quantization steps $M$.
To test whether \OUR benefits from having $M$ different neural networks $f_{\theta_m}$, we share (a subset of the) parameters among the $M$ steps and observed drops in performance. Yet, performance remained superior to LSQ in all tested cases.

\section{Conclusion}

We introduced \OUR, a neural vector quantizer based on residual quantization. \OUR has the unique property that it adapts the codebook for each quantization step to the distribution of residual vectors in the current quantization cell. To achieve this, \OUR leverages a neural network that is conditioned upon the selected codewords in previous steps, and generates a specialized codebook for the next step. The implicitly-available set of available codebooks grows exponentially with the number of quantization steps, which makes \OUR a very flexible multi-codebook quantizer.
We experimentally validate \OUR and compare it to state-of-the-art baselines on six different datasets. We observe substantial improvements in quantization performance, as measured by the reconstruction error, and nearest-neighbor search accuracy. 
We show that \OUR can be combined with inverted file indexing for efficient large-scale vector search, and that this reaches new high-accuracy operating points.
Finally, we find that truncating \OUR codes during encoding or decoding, results in quantization performance that is on par with \OUR models trained for smaller bit rates. 
This makes \OUR an effective multi-rate quantizer.

\OUR opens several directions for further research, \eg to explore implicit neural codebooks for other quantization schemes such as product quantization, in designs specifically tailored to fast nearest-neighbor search, and for compression of media such as audio, images or videos. On the algorithmic level, we plan to explore the use of beam search during \OUR encoding in future work to investigate whether a possible improvement in accuracy outweighs the added complexity.

\section*{Impact Statement}
\label{sec:impact_statement}

This paper presents work whose goal is to advance the state of the art in data compression and similarity search. 
Although there are many potential societal consequences of our work, we feel none of them must be specifically highlighted here as our contributions do not enable specific new use cases but rather improve existing ones.

\bibliographystyle{icml2024}
\bibliography{bibliography}

\begin{thebibliography}{56}
\providecommand{\natexlab}[1]{#1}
\providecommand{\url}[1]{\texttt{#1}}
\expandafter\ifx\csname urlstyle\endcsname\relax
  \providecommand{\doi}[1]{doi: #1}\else
  \providecommand{\doi}{doi: \begingroup \urlstyle{rm}\Url}\fi

\bibitem[Agustsson et~al.(2017)Agustsson, Mentzer, Tschannen, Cavigelli, Benini, and Van~Gool]{Agustsson2017}
Agustsson, E., Mentzer, F., Tschannen, M., Cavigelli, L., Benini, L., and Van~Gool, L.
\newblock {Soft-to-hard vector quantization for end-to-end learning compressible representations}.
\newblock In \emph{NeurIPS}, 2017.

\bibitem[Amara et~al.(2022)Amara, Douze, Sablayrolles, and J{\'{e}}gou]{Amara2022NearestPerspective}
Amara, K., Douze, M., Sablayrolles, A., and J{\'{e}}gou, H.
\newblock Nearest neighbor search with compact codes: A decoder perspective.
\newblock In \emph{ICMR}, 2022.

\bibitem[Andrychowicz et~al.(2016)Andrychowicz, Denil, Gómez, Hoffman, Pfau, Schaul, Shillingford, and de~Freitas]{andrychowicz16nips}
Andrychowicz, M., Denil, M., Gómez, S., Hoffman, M.~W., Pfau, D., Schaul, T., Shillingford, B., and de~Freitas, N.
\newblock Learning to learn by gradient descent by gradient descent.
\newblock In \emph{NeurIPS}, 2016.

\bibitem[Babenko \& Lempitsky(2014)Babenko and Lempitsky]{babenko2014additive}
Babenko, A. and Lempitsky, V.
\newblock Additive quantization for extreme vector compression.
\newblock In \emph{CVPR}, 2014.

\bibitem[Babenko \& Lempitsky(2015)Babenko and Lempitsky]{Babenko2015TreeClassification}
Babenko, A. and Lempitsky, V.
\newblock {Tree quantization for large-scale similarity search and classification}.
\newblock In \emph{CVPR}, 2015.

\bibitem[Babenko \& Lempitsky(2016)Babenko and Lempitsky]{babenko2016efficient}
Babenko, A. and Lempitsky, V.
\newblock Efficient indexing of billion-scale datasets of deep descriptors.
\newblock In \emph{CVPR}, 2016.

\bibitem[Baranchuk et~al.(2018)Baranchuk, Babenko, and Malkov]{baranchuk2018revisiting}
Baranchuk, D., Babenko, A., and Malkov, Y.
\newblock Revisiting the inverted indices for billion-scale approximate nearest neighbors.
\newblock In \emph{ECCV}, 2018.

\bibitem[Chang et~al.(2022)Chang, Zhang, Jiang, Liu, and Freeman]{chang22cvpr}
Chang, H., Zhang, H., Jiang, L., Liu, C., and Freeman, W.~T.
\newblock {MaskGIT}: Masked generative image transformer.
\newblock In \emph{CVPR}, 2022.

\bibitem[Chen et~al.(2010)Chen, Guan, and Wang]{chen2010approximate}
Chen, Y., Guan, T., and Wang, C.
\newblock {Approximate nearest neighbor search by residual vector quantization}.
\newblock \emph{Sensors}, 10\penalty0 (12):\penalty0 11259--11273, 2010.

\bibitem[Copet et~al.(2023)Copet, Kreuk, Gat, Remez, Kant, Synnaeve, Adi, and D{\'e}fossez]{copet2023simple}
Copet, J., Kreuk, F., Gat, I., Remez, T., Kant, D., Synnaeve, G., Adi, Y., and D{\'e}fossez, A.
\newblock Simple and controllable music generation.
\newblock In \emph{NeurIPS}, 2023.

\bibitem[Cover \& Thomas(1991)Cover and Thomas]{cover91eit}
Cover, T.~M. and Thomas, J.~A.
\newblock \emph{{Elements of Information Theory}}.
\newblock John Wiley \& Sons, 1991.

\bibitem[D{\'e}fossez et~al.(2023)D{\'e}fossez, Copet, Synnaeve, and Adi]{defossez2023high}
D{\'e}fossez, A., Copet, J., Synnaeve, G., and Adi, Y.
\newblock {High fidelity neural audio compression}.
\newblock \emph{Transactions on Machine Learning Research}, 2023.

\bibitem[Devlin et~al.(2018)Devlin, Chang, Lee, and Toutanova]{devlin2018bert}
Devlin, J., Chang, M.-W., Lee, K., and Toutanova, K.
\newblock {BERT: Pre-training of deep bidirectional transformers for language understanding}.
\newblock In \emph{Proceedings of North American Chapter of the Association for Computational Linguistics (NAACL)}, 2018.

\bibitem[Douze et~al.(2016)Douze, J{\'e}gou, and Perronnin]{douze2016polysemous}
Douze, M., J{\'e}gou, H., and Perronnin, F.
\newblock Polysemous codes.
\newblock In \emph{ECCV}, 2016.

\bibitem[Douze et~al.(2024)Douze, Guzhva, Deng, Johnson, Szilvasy, Mazar{\'e}, Lomeli, Hosseini, and J{\'e}gou]{douze2024faiss}
Douze, M., Guzhva, A., Deng, C., Johnson, J., Szilvasy, G., Mazar{\'e}, P.-E., Lomeli, M., Hosseini, L., and J{\'e}gou, H.
\newblock The {Faiss} library.
\newblock \emph{arXiv preprint}, 2401.08281, 2024.

\bibitem[El-Nouby et~al.(2023)El-Nouby, Muckley, Ullrich, Laptev, Verbeek, and J{\'{e}}gou]{El-Nouby2022}
El-Nouby, A., Muckley, M.~J., Ullrich, K., Laptev, I., Verbeek, J., and J{\'{e}}gou, H.
\newblock Image compression with product quantized masked image modeling.
\newblock \emph{Transactions on Machine Learning Research}, 2023.

\bibitem[Esser et~al.(2021)Esser, Rombach, and Ommer]{esser2021taming}
Esser, P., Rombach, R., and Ommer, B.
\newblock Taming transformers for high-resolution image synthesis.
\newblock In \emph{CVPR}, 2021.

\bibitem[Ge et~al.(2013)Ge, He, Ke, and Sun]{ge2013optimized}
Ge, T., He, K., Ke, Q., and Sun, J.
\newblock Optimized product quantization for approximate nearest neighbor search.
\newblock In \emph{CVPR}, 2013.

\bibitem[Glynn(1990)]{glynn1990likelihood}
Glynn, P.~W.
\newblock Likelihood ratio gradient estimation for stochastic systems.
\newblock \emph{Communications of the ACM}, 33\penalty0 (10):\penalty0 75--84, 1990.

\bibitem[Gray(1984)]{gray1984vector}
Gray, R.
\newblock {Vector quantization}.
\newblock \emph{{\sc IEEE} Transactions on Acoustics, Speech and Signal Processing}, 1\penalty0 (2):\penalty0 4--29, 1984.

\bibitem[Guo et~al.(2020)Guo, Sun, Lindgren, Geng, Simcha, Chern, and Kumar]{guo2020accelerating}
Guo, R., Sun, P., Lindgren, E., Geng, Q., Simcha, D., Chern, F., and Kumar, S.
\newblock Accelerating large-scale inference with anisotropic vector quantization.
\newblock In \emph{ICML}, 2020.

\bibitem[He et~al.(2013)He, Wen, and Sun]{he2013kmeanshashing}
He, K., Wen, F., and Sun, J.
\newblock K-means hashing: An affinity-preserving quantization method for learning binary compact codes.
\newblock In \emph{CVPR}, 2013.

\bibitem[He et~al.(2016)He, Zhang, Ren, and Sun]{he16cvpr}
He, K., Zhang, X., Ren, S., and Sun, J.
\newblock Deep residual learning for image recognition.
\newblock In \emph{CVPR}, 2016.

\bibitem[He et~al.(2023)He, Gao, Song, and Li]{he2023semisupervised}
He, T., Gao, L., Song, J., and Li, Y.-F.
\newblock Semisupervised network embedding with differentiable deep quantization.
\newblock \emph{{\sc IEEE} Transactions on Neural Networks and Learning Systems}, 34\penalty0 (8):\penalty0 4791--4802, 2023.

\bibitem[Huijben et~al.(2022)Huijben, Kool, Paulus, and Van~Sloun]{huijben2022review}
Huijben, I.~A., Kool, W., Paulus, M.~B., and Van~Sloun, R.~J.
\newblock A review of the {Gumbel}-max trick and its extensions for discrete stochasticity in machine learning.
\newblock \emph{{\sc IEEE} Transactions on Pattern Analysis and Machine Intelligence}, 45\penalty0 (2):\penalty0 1353--1371, 2022.

\bibitem[Izacard et~al.(2022)Izacard, Caron, Hosseini, Riedel, Bojanowski, Joulin, and Grave]{izacard2021contriever}
Izacard, G., Caron, M., Hosseini, L., Riedel, S., Bojanowski, P., Joulin, A., and Grave, E.
\newblock Unsupervised dense information retrieval with contrastive learning.
\newblock \emph{Transactions on Machine Learning Research}, 2022.

\bibitem[Jang et~al.(2017)Jang, Gu, and Poole]{Jang2017}
Jang, E., Gu, S., and Poole, B.
\newblock Categorical reparameterization with {Gumbel-Softmax}.
\newblock In \emph{ICLR}, 2017.

\bibitem[Jang \& Cho(2021)Jang and Cho]{jang2021self}
Jang, Y.~K. and Cho, N.~I.
\newblock Self-supervised product quantization for deep unsupervised image retrieval.
\newblock In \emph{ICCV}, 2021.

\bibitem[J\'egou et~al.(2010)J\'egou, Douze, and Schmid]{jegou2010product}
J\'egou, H., Douze, M., and Schmid, C.
\newblock Product quantization for nearest neighbor search.
\newblock \emph{{\sc IEEE} Transactions on Pattern Analysis and Machine Intelligence}, 33\penalty0 (1):\penalty0 117--128, 2010.

\bibitem[J{\'e}gou et~al.(2011)J{\'e}gou, Tavenard, Douze, and Amsaleg]{jegou2011searching}
J{\'e}gou, H., Tavenard, R., Douze, M., and Amsaleg, L.
\newblock Searching in one billion vectors: Re-rank with source coding.
\newblock In \emph{ICASSP}, 2011.

\bibitem[Kingma \& Ba(2015)Kingma and Ba]{kingma2014adam}
Kingma, D.~P. and Ba, J.
\newblock Adam: A method for stochastic optimization.
\newblock In \emph{ICLR}, 2015.

\bibitem[Klein \& Wolf(2019)Klein and Wolf]{klein2019end}
Klein, B. and Wolf, L.
\newblock End-to-end supervised product quantization for image search and retrieval.
\newblock In \emph{CVPR}, 2019.

\bibitem[Kumar et~al.(2023)Kumar, Seetharaman, Luebs, Kumar, and Kumar]{kumar2023high}
Kumar, R., Seetharaman, P., Luebs, A., Kumar, I., and Kumar, K.
\newblock High-fidelity audio compression with improved {RVQGAN}.
\newblock In \emph{NeurIPS}, 2023.

\bibitem[Lee et~al.(2022)Lee, Kim, Kim, Cho, and Han]{Lee_2022_CVPR}
Lee, D., Kim, C., Kim, S., Cho, M., and Han, W.-S.
\newblock Autoregressive image generation using residual quantization.
\newblock In \emph{CVPR}, 2022.

\bibitem[Liu et~al.(2018)Liu, Cao, Long, Wang, and Wang]{liu2018deep}
Liu, B., Cao, Y., Long, M., Wang, J., and Wang, J.
\newblock Deep triplet quantization.
\newblock In \emph{ACM International conference on Multimedia}, 2018.

\bibitem[Ma et~al.(2020)Ma, Zhang, Huang, and Sun]{ma2020weightnet}
Ma, N., Zhang, X., Huang, J., and Sun, J.
\newblock Weightnet: Revisiting the design space of weight networks.
\newblock In \emph{ECCV}, 2020.

\bibitem[Maddison et~al.(2017)Maddison, Mnih, and Teh]{Maddison2017}
Maddison, C.~J., Mnih, A., and Teh, Y.~W.
\newblock The concrete distribution: A continuous relaxation of discrete random variables.
\newblock In \emph{ICLR}, 2017.

\bibitem[Malkov \& Yashunin(2018)Malkov and Yashunin]{malkov2018hnsw}
Malkov, Y.~A. and Yashunin, D.~A.
\newblock Efficient and robust approximate nearest neighbor search using hierarchical navigable small world graphs.
\newblock \emph{{\sc IEEE} Transactions on Pattern Analysis and Machine Intelligence}, 42\penalty0 (4):\penalty0 824--836, 2018.

\bibitem[Martinez et~al.(2016)Martinez, Clement, Hoos, and Little]{martinez2016revisiting}
Martinez, J., Clement, J., Hoos, H.~H., and Little, J.~J.
\newblock Revisiting additive quantization.
\newblock In \emph{ECCV}, 2016.

\bibitem[Martinez et~al.(2018)Martinez, Zakhmi, Hoos, and Little]{martinez2018lsq++}
Martinez, J., Zakhmi, S., Hoos, H.~H., and Little, J.~J.
\newblock {LSQ++}: Lower running time and higher recall in multi-codebook quantization.
\newblock In \emph{ECCV}, 2018.

\bibitem[Morozov \& Babenko(2019)Morozov and Babenko]{Morozov2019UnsupervisedSearch}
Morozov, S. and Babenko, A.
\newblock Unsupervised neural quantization for compressed-domain similarity search.
\newblock In \emph{ICCV}, 2019.

\bibitem[Niu et~al.(2023)Niu, Xu, Zhao, He, Ji, Yuan, and Xue]{Niu2023ResidualSearch}
Niu, L., Xu, Z., Zhao, L., He, D., Ji, J., Yuan, X., and Xue, M.
\newblock Residual vector product quantization for approximate nearest neighbor search.
\newblock \emph{Expert Systems with Applications}, 232, 2023.

\bibitem[Noh et~al.(2023)Noh, Hyun, Jeong, Lim, and Heo]{noh2023disentangled}
Noh, H., Hyun, S., Jeong, W., Lim, H., and Heo, J.-P.
\newblock Disentangled representation learning for unsupervised neural quantization.
\newblock In \emph{CVPR}, 2023.

\bibitem[Oquab et~al.(2023)Oquab, Darcet, Moutakanni, Vo, Szafraniec, Khalidov, Fernandez, Haziza, Massa, El-Nouby, et~al.]{oquab2023dinov2}
Oquab, M., Darcet, T., Moutakanni, T., Vo, H., Szafraniec, M., Khalidov, V., Fernandez, P., Haziza, D., Massa, F., El-Nouby, A., et~al.
\newblock {DINOv2: Learning Robust Visual Features Without Supervision}.
\newblock \emph{Transactions on Machine Learning Research}, 2023.

\bibitem[Paterek(2007)]{paterek2007improving}
Paterek, A.
\newblock Improving regularized singular value decomposition for collaborative filtering.
\newblock In \emph{Proceedings of KDD cup and workshop}, 2007.

\bibitem[Pizzi et~al.(2022)Pizzi, Roy, Ravindra, Goyal, and Douze]{pizzi2022self}
Pizzi, E., Roy, S.~D., Ravindra, S.~N., Goyal, P., and Douze, M.
\newblock A self-supervised descriptor for image copy detection.
\newblock In \emph{CVPR}, 2022.

\bibitem[Radford et~al.(2021)Radford, Kim, Hallacy, Ramesh, Goh, Agarwal, Sastry, Askell, Mishkin, Clark, Krueger, and Sutskever]{radford21clip}
Radford, A., Kim, J.~W., Hallacy, C., Ramesh, A., Goh, G., Agarwal, S., Sastry, G., Askell, A., Mishkin, P., Clark, J., Krueger, G., and Sutskever, I.
\newblock Learning transferable visual models from natural language supervision.
\newblock In \emph{ICML}, 2021.

\bibitem[Schwenk \& Douze(2017)Schwenk and Douze]{schwenk2017learning}
Schwenk, H. and Douze, M.
\newblock Learning joint multilingual sentence representations with neural machine translation.
\newblock In \emph{Workshop on Representation Learning for NLP}, 2017.

\bibitem[Simhadri et~al.(2022)Simhadri, Williams, Aum{\"u}ller, Douze, Babenko, Baranchuk, Chen, Hosseini, Krishnaswamny, Srinivasa, et~al.]{simhadri2022results}
Simhadri, H.~V., Williams, G., Aum{\"u}ller, M., Douze, M., Babenko, A., Baranchuk, D., Chen, Q., Hosseini, L., Krishnaswamny, R., Srinivasa, G., et~al.
\newblock Results of the {NeurIPS’21} challenge on billion-scale approximate nearest neighbor search.
\newblock In \emph{NeurIPS 2021 Competitions and Demonstrations Track}, 2022.

\bibitem[Subramanya et~al.(2019)Subramanya, Kadekodi, Krishaswamy, and Simhadri]{subramanya2019diskann}
Subramanya, S.~J., Kadekodi, R., Krishaswamy, R., and Simhadri, H.~V.
\newblock {DiskANN}: Fast accurate billion-point nearest neighbor search on a single node.
\newblock In \emph{NeurIPS}, 2019.

\bibitem[van~den Oord et~al.(2017)van~den Oord, Vinyals, and Kavukcuoglu]{VandenOordDeepMind}
van~den Oord, A., Vinyals, O., and Kavukcuoglu, K.
\newblock Neural discrete representation learning.
\newblock In \emph{NeurIPS}, 2017.

\bibitem[Wang et~al.(2022)Wang, Zeng, Chen, Dai, and Xia]{wang2022contrastive}
Wang, J., Zeng, Z., Chen, B., Dai, T., and Xia, S.-T.
\newblock Contrastive quantization with code memory for unsupervised image retrieval.
\newblock In \emph{AAAI}, 2022.

\bibitem[Williams(1992)]{williams1992simple}
Williams, R.~J.
\newblock Simple statistical gradient-following algorithms for connectionist reinforcement learning.
\newblock \emph{Machine learning}, 8\penalty0 (3):\penalty0 229--256, 1992.

\bibitem[Ypsilantis et~al.(2023)Ypsilantis, Chen, Cao, Lipovsk{\`y}, Dogan-Sch{\"o}nberger, Makosa, Bluntschli, Seyedhosseini, Chum, and Araujo]{ypsilantis2023towards}
Ypsilantis, N.-A., Chen, K., Cao, B., Lipovsk{\`y}, M., Dogan-Sch{\"o}nberger, P., Makosa, G., Bluntschli, B., Seyedhosseini, M., Chum, O., and Araujo, A.
\newblock Towards universal image embeddings: A large-scale dataset and challenge for generic image representations.
\newblock In \emph{ICCV}, 2023.

\bibitem[Yu et~al.(2018)Yu, Yuan, Fang, and Jin]{yu2018product}
Yu, T., Yuan, J., Fang, C., and Jin, H.
\newblock Product quantization network for fast image retrieval.
\newblock In \emph{ECCV}, 2018.

\bibitem[Zhu et~al.(2023)Zhu, Song, Gao, Gu, and Shen]{Zhu2023RevisitingQuantization}
Zhu, X., Song, J., Gao, L., Gu, X., and Shen, H.~T.
\newblock Revisiting multi-codebook quantization.
\newblock \emph{{\sc IEEE} Transactions on Image Processing}, 32:\penalty0 2399--2412, 2023.

\end{thebibliography}
\newpage
\appendix
\onecolumn

\clearpage

\setcounter{figure}{0} 
\renewcommand\thefigure{S\arabic{figure}} 
\setcounter{table}{0}
\renewcommand{\thetable}{S\arabic{table}}

\section{Implementation details}
\label{app:implementation_details}

\subsection{IVF Faiss implementation}
\label{app:ivfimplem}

Faiss has a residual quantization implementation combined with an inverted file (IVF-RQ). 
The corresponding index factory name that we use for the 16-byte experiments is \verb|IVF1048576_HNSW32,RQ16x8_Nqint8|, which gives the number of IVF centroids ($\nlist=2^{20}$), indexed with a HNSW graph-based index (32 links per node), the size of the RQ ($16\times8$ bits) and how the norm is encoded for fast search (with an 8-bit integer). 
To build the IVF-RQ we also set the beam size directly in the index. 
The 1M IVF centroids are obtained by running k-means on GPU, but otherwise the IVF-RQ experiments run only on CPU, as IVF-RQ is not implementated on GPU in Faiss. 

It turns out that this index structure can be used as-is for the \IVFOUR experiments because the decoder and fast-search functionality of IVF-RQ and \IVFOUR are the same: both are an AQ decoder.
Therefore, we build an IVF-RQ index, set the codebook tables to $\vG$ (\cref{sec:LSdecoder}) and fill in the index with pre-computed \OUR codes for the databse vectors.

At search time, the Faiss index is used to retrieve the top-$\nshort$ search results and the corresponding codes (that are extracted from the inverted lists). 
The decoding and re-ranking is performed in Pytorch. 
The total search time is thus the sum of 
(1) the initial search time (that depends on $\nprobe$ and \texttt{efSearch}), 
(2) the \OUR decoding time (that depends on $\nshort$) and 
(3) the distance computations and reranking (that are normally very fast).

\subsection{Training UNQ}
\label{app:experiments_with_UNQ}
We use the author's code of UNQ~\citep{Morozov2019UnsupervisedSearch} to replicate their experimental results and run additional experiments. 
We noticed that the original code picks the best model based on R@1 accuracy on the query set that was also used to report results, which is overly optimistic for real-world settings. 
To correct for this, we use the same validation set as in the \OUR experiments, but exploited those vectors as validation queries and picked the best model based on R@1 performance of those. 
As such, for our UNQ reproductions, recall numbers may be slightly lower than reported in the original paper~\citep{Morozov2019UnsupervisedSearch}.

We wanted to test the scalability of UNQ, both in terms of model capacity and number of training vectors. 
However, UNQ's triplet loss requires substantial compute for mining negative samples, as it does a nearest-neighbor search of all vectors in the training set, each time a new set of negatives needs to be drawn. 
Running this search is feasible on 500k training vectors, as used in the experiments reported in the original UNQ paper, but for 10M vectors it results in infeasible running times where a single negative mining pass takes over eight hours.
However, as noted by the UNQ authors in an ablation of their paper~\citep[Table 5]{Morozov2019UnsupervisedSearch}, the triplet loss term does not contribute substantially, and actually decreases performance for R@1 and R@10 for the tested setting (BigANN1M, 8 bytes). 
As such, we set $\alpha=0$ in~\citep[Eq. 12]{Morozov2019UnsupervisedSearch} when running UNQ on 10M vectors, which turns off the triplet loss. 
This enables scaling experiments to 10M training vectors. 
UNQ* models in \cref{tab:scaling_training_vecs_FAISSbaselines} and all results in \cref{fig:scaling_unq} are trained as described above.

A final challenge we faced when training UNQ was instability.
When increasing the capacity (either by increasing the width or depth of the encoder/decoder), the training gets stuck due to large gradients when the learning rate is set to $10^{-3}$ as proposed by the authors. 
For this reason, we also experimented with a learning rate of $10^{-4}$, which stabilized a substantial portion of the runs.
For all UNQ experiments reported in this supplemental material, we tested both learning rates ($10^{-3}$ and $10^{-4}$), and report the best performing UNQ model.

\subsection{Training \bf \OUR}
\label{app:training_details}

\OUR and its variants were implemented in Pytorch 2.0.1 and trained using the Adam optimizer with default settings~\citep{kingma2014adam} across eight GPUs with an effective batch size of 1,024. %
The same seed for randomization was used in all experiments. 
The base learning rate was reduced by a factor 10 every time the loss on the validation set did not improve for 10 epochs.
We stopped training when the validation loss did not improve for 50 epochs. In general this happened within 200--350 epochs, depending on the model size and dataset.

During training, we compute the loss from \cref{eq:loss} in two passes: (1) an encoding of the training batch without tracking the gradients, and (2) computation of the loss with gradients when the codes are known. This speeds up the computation 2.5$\times$ compared to a naive implementation.

When we trained \OUR on the small training set (\ie T=500k) we noticed that for some datasets, a base learning rate of $10^{-3}$ resulted in slightly better performance than a base rate of $10^{-4}$. 
However for some of the larger \OUR models trained on 10M vectors a lower base learning rate worked better. 
We opted for a uniform setting of $10^{-4}$ that can be used in all models and datasets, 
$10^{-3}$ was only used when mentioned explicitly in the text.

To initialize the base codebooks $\bar{\vC}$, we used the RQ implementation from the Faiss library~\citep{douze2024faiss}, with a beam size $B=1$. This resulted in competitive or slightly better performance than the default $B=5$, presumably because for \OUR we also used a greedy assignment (equivalent to a beam size of one).

\section{Additional analyses}
\label{app:additional_analyzes}

\subsection{Capacity of \bf \OUR}
\label{app:L_vs_h}

\begin{figure}
    \centering
    \includegraphics[width=.5\columnwidth]{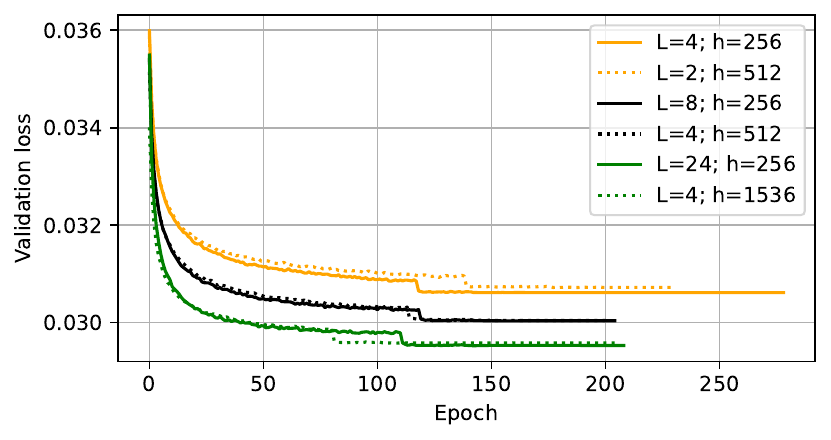}
    \caption{Validation loss on 8 bytes encoding \OUR models trained on 10M BigANN. Changing the model capacity using either $L$ or $h$ with the same factor similarly affects  validation loss.}
    \label{fig:L_vs_h}
\end{figure}

The number of trainable parameters scales linearly with both the number of residual blocks $L$ and the hidden dimension $h$ of the residual-MLPs, see \cref{eq:nr_parameters}. 
\Cref{fig:L_vs_h} plots the validation loss of different 8-bytes \OUR models trained on BigANN. 
Curves with the same color have the same model capacity, but differ in $L$ and $h$. It can be seen that changing one or the other has a similar effect on model performance. A slight advantage is visible for increasing $L$ rather than $h$. For that reason ---in order to create only one parameter that influences model capacity--- we propose to fix $h\!=\!256$ and adjust $L$ to change the capacity of \OUR.

\begin{table}[t]
    \centering
    \caption{Entropy $\mathcal{H}$ of codeword assignments, averaged over codebooks, of the compressed database.
}
    {\scriptsize
  \begin{tabular}{clHHHHcHHHHcHHHHHHHHHH}
    \toprule
         && 
         \multicolumn{5}{c}{\bf BigANN1M}&  \multicolumn{5}{c}{\bf Deep1M} \\
          \midrule
 \multirow{5}{*}{\rotatebox{90}{\bf8 bytes}} & 
    OPQ& 2.95  & 21.9 & 64.8 & 95.4 & 7.90 &  0.26 & 15.9 & 51.2 & 88.2  & 7.95 & 1.87 & 8.0 & 24.7 & 50.8 & 7.78 & 9.52 & 2.5 & 5.1 & 10.9 & 8.00  \\
         &RQ& 2.49 & 27.9 & 75.2 & 98.2 & 7.95 & 0.20 & 21.4 & 63.5 & 95.2 & 7.96 & 1.82 & 10.2 & 26.9 & 52.4 & 7.90 &  9.20 & 2.7 & 6.1 & 13.6 & 7.99\\
         &LSQ& 1.91 & 31.9 & 79.5 & 98.9 & 7.95 & 0.17 & 24.6 & 69.4 & 97.0 & 7.95 &  1.65 & 13.1  & 33.9 & 62.7 & 7.92 & 8.87& 3.3 & 7.5 & 17.3 & 7.99 \\
         &UNQ  & 1.51 & 34.6 & 82.8 & 99.0 & 8.00 & 0.16 & 26.7 & 72.6 & 97.3 & 7.99 & --- &  --- &  --- &  --- &  --- &   --- &  --- &  --- &  --- &  --- \\
         &\OUR & \bf 1.12  & \bf 45.2 & \bf 91.2 &\bf  99.7 & 7.99 & \bf 0.12 & \bf 36.3 & \bf 84.6 & \bf 99.4 & 7.99 & \bf 1.40 & \bf 20.7 & \bf 47.4 & \bf 74.6 & 7.91 & \bf 8.67 & \bf 3.6 & \bf 8.9 & \bf 20.6 & 8.00\\
         \midrule
         \multirow{5}{*}{\rotatebox{90}{\bf16 bytes}} &
         OPQ& 1.79 & 40.5 & 89.9 & 99.8 & 7.94 & 0.14 & 34.9 & 82.2 & 98.9 & 7.93 & 1.71 & 18.3 & 40.9 & 65.4 & 7.95 & 7.25 & 5.0 & 11.8 & 25.9 & 8.00\\
         &RQ& 1.30 & 49.0 & 95.0 & 100.0 & 7.97 & 0.10 & 43.0 & 90.8 & 99.8 & 7.98 & 1.65 & 20.2 & 43.5 & 68.2 & 7.95 & 7.01 & 5.4 & 13.0 & 29.0 & 8.00\\
         &LSQ& 0.98 & 51.1 & 95.4 & 100.0 & 7.93 & 0.09 & 42.3 & 89.7 & 99.8 & 7.94 &1.35 & 25.6 & 53.8 & 78.6 & 7.91 & 6.63 & 6.2 & 14.8 & 32.3 & 8.00\\
         &UNQ  & 0.57 & 59.3 & 98.0 & 100.0 & 7.99 & 0.07 & 47.9 & 93.0 & 99.8 & 7.99 &  --- &  --- &  --- &  --- &  --- &  --- &  --- &  --- &  --- &  --- \\
         &\OUR &\bf 0.32 &\bf 71.9 & \bf 99.6 & \bf 100.0 & 7.99 & \bf 0.05 & \bf 59.8 & \bf 98.0 & \bf 100.0 & 7.99 & {\bf 1.10}  & {\bf 31.1}  & \bf 62.0&  \bf 85.9  & X  & \bf6.58 & \bf 6.4 & \bf 16.8 & \bf 35.5 & 8.00 \\
          \bottomrule
    \end{tabular}
    }
    \label{tab:entropies}
\end{table}

\subsection{Codeword usage}
\label{app:entropy}

To investigate whether \OUR suffers from codebook collapse --- a common problem in neural quantization models --- one can use the average Shannon entropy (averaged over codebooks) to expresses the distribution of selected codewords by the compressed database. It is defined as: 
$\mathcal{H} = -\frac{1}{M}\sum_{m=1}^{M}\sum_{k=1}^{K} p^m_k \log_2(p^m_k), $
and upper-bounded by $\log_2(K)$ bits. 
Here, $p^m_k$ is the empirical probability that the $k^\text{th}$ codeword gets assigned in the $m^\text{th}$ codebook when compressing the full database. 

We find that \OUR achieves near-optimal codeword usage, $\mathcal{H} \approx \log_2(K)$ bits, in all cases, see \cref{tab:entropies}. 
Note that UNQ~\citep{Morozov2019UnsupervisedSearch} also achieves this, but it requires regularization at training time, which introduces an additional hyperparameter that weighs this regularizing term. Also the authors of DeepQ~\citep{Zhu2023RevisitingQuantization} propose to use such a regularization term. 

The fact that \OUR is not reliant on such additional regularization can be attributed to 
(i) \OUR is initialized with base codebooks using RQ that enforces a good initial spread of assignments, and 
(ii) since \OUR does not deploy an encoder before quantization, codebook collapse by the encoder, where all data vectors are mapped to a similar point in latent space, cannot occur.

\begin{figure*}[t]
    \centering
    \includegraphics[width=\columnwidth]{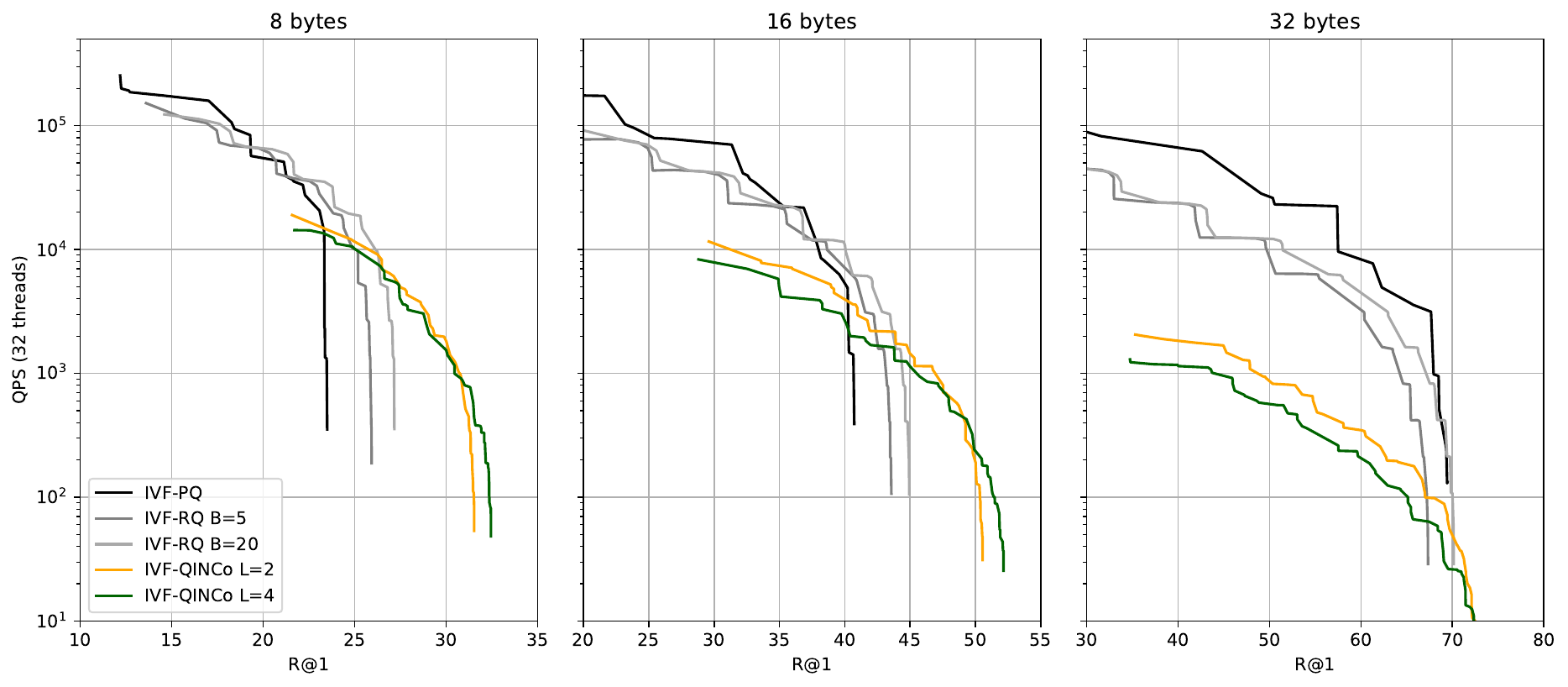}
    \caption{
        Speed in queries per second (QPS) vs search accuracy (R@1) trade-offs for the Deep1B dataset.
    }
    \label{fig:ivfdeep}
\end{figure*}

\subsection{Fast search}
\label{app:fast_search}

\mypar{Results on Deep1B}
\Cref{fig:ivfdeep} shows the speed-recall trade-offs for the Deep1B dataset, similar to the results shown for  BigANN1B  in \cref{fig:IVFplot} of the main paper.
There is a wide range of high-accuracy operating points where QINCo is competitive or outperforms IVF-PQ and IVF-RQ for 8 and 16-byte encoding. 
The trade-offs for the 32-byte setting are less interesting %
compared to RQ and PQ, because here the upper bound accuracy of \OUR \wrt these methods is not high enough.
It is possible that PQ-\OUR would be a better option in this case. 

Both for BigANN1B (\cref{fig:IVFplot}) and Deep1B (\cref{fig:ivfdeep}), it can be seen that the capacity parameter $L$ slightly changes the Pareto front (green \vs yellow curves). 
At high accuracy operating points, \IVFOUR with $L\!=\!2$ starts to become slower than \IVFOUR with $L\!=\!4$, which seems counter-intuitive. 
This, however, is caused by the fact that in this regime, \IVFOUR with $L\!=\!2$ requires a longer short-list (higher $\nshort$) than  \IVFOUR with $L\!=\!4$ to achieve the same accuracy, while at lower accuracies \IVFOUR with $L\!=\!2$ is faster due to its lower decoding complexity.

\begin{figure}
    \centering
    \includegraphics[width=.45\columnwidth]{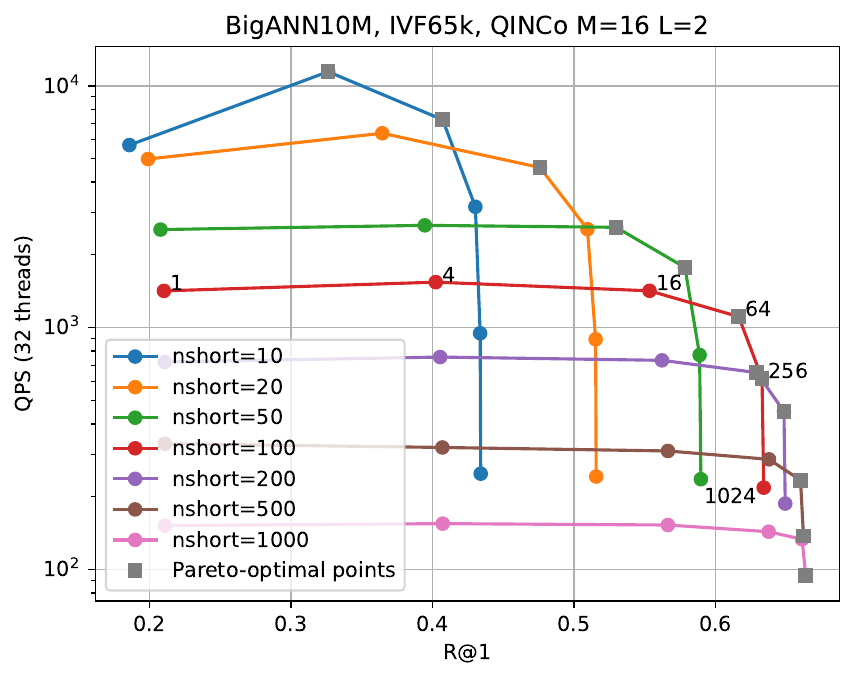}
    \caption{
        All combinations of $\nprobe$ and $\nshort$ for one dataset.
        For some points we indicate the $\nprobe$ value. 
    }
    \label{fig:bigann10M}
\end{figure}

\mypar{Decomposing performance over parameters}
Pareto-optimal curves do not show the runtime parameters that are used in each experiment. 
\Cref{fig:bigann10M} shows all the combination of parameters for a small experiment with 10M database elements and an IVF index of just $2^{16}=64$k centroids. 
In this case, the IVF centroids are searched exhaustively, without an approximate HNSW index, so there is no \texttt{efSearch} parameter involved. 
This makes it possible to show all parameter combinations. 
The Pareto-optimal points are indicated in gray squares: they are the ones that give the best accuracy for a given time budget or conversely the fastest search for a given recall requirement.

\begin{figure*}
    \centering
    \includegraphics[width=\textwidth]{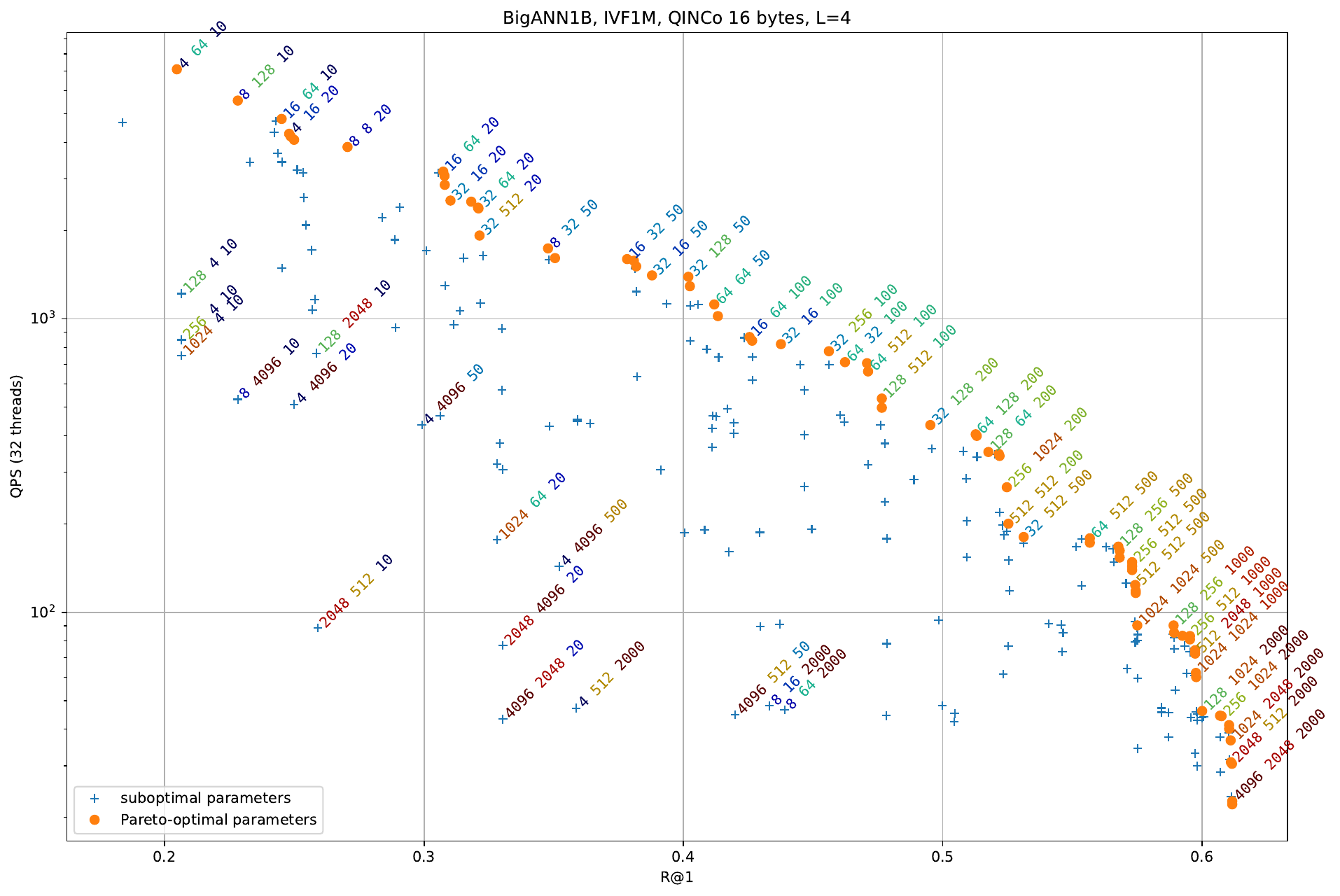}
    \caption{
        The set of parameters that are tried out for one of the curves of \cref{fig:IVFplot}. 
        Each point is obtained by setting three parameters: the $\nprobe$, HNSW's \texttt{efSearch} and $\nshort$. 
        We indicate the values of these parameters (in this order) for some of the results and color them from lowest (blue) to highest (red) with green in-between.
    }
    \label{fig:bigann1Bannotated}
\end{figure*}

\Cref{fig:bigann1Bannotated} show the same trade-offs for the BigANN1B dataset for a subset of the parameter sets. 
It shows that for Pareto-optimal points, the three considered parameters need to be set to ``compatible'' values: it is useless to set a high $\nprobe$ with a low $\nshort$ and vice-versa. 
The granularity of the parameter we tried out is relatively coarse. 
The settings for $\nprobe$ are clearly separated and there are probably slightly better operating points for intermediate settings like $\nshort=30$ or $\nshort=700$.

\subsection{Scaling baselines}
\label{app:scaling_unq}

\Cref{tab:scaling_training_vecs_FAISSbaselines} shows the performance for \OUR and all baselines both trained on 500k vectors and 10M vectors. OPQ, RQ and LSQ do not benefit from more training data in general, while UNQ did improve. A more detailed analysis on UNQ's scalability follows in this section.

In \cref{tab:scaling_training_vecs_FAISSbaselines}, for 500k training vectors we use the original numbers from the paper~\citep{Morozov2019UnsupervisedSearch}, while we denote with UNQ$^*$ results we obtained by training on 10M vectors  by re-running the author's codebase, while model selection was based on the hold-out validation set
that we created,  see \cref{app:experiments_with_UNQ}. 
    The triplet loss was not used in this scenario as the negative mining on 10M training vectors resulted in prohibitively slow training. 
    
\begin{figure*}
    \centering
    \includegraphics[width=.4\columnwidth]{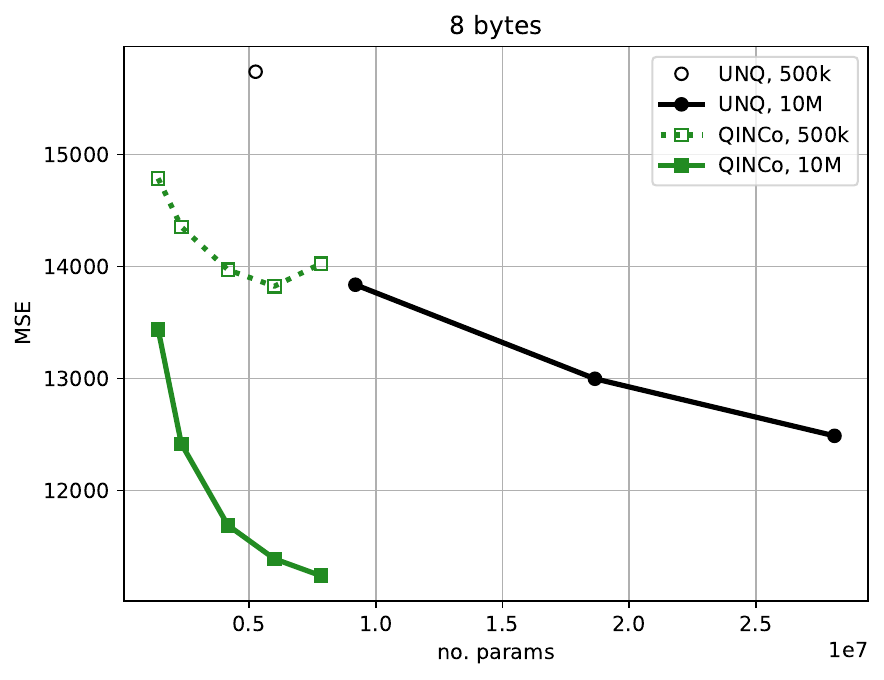}
    \includegraphics[width=.4\columnwidth]{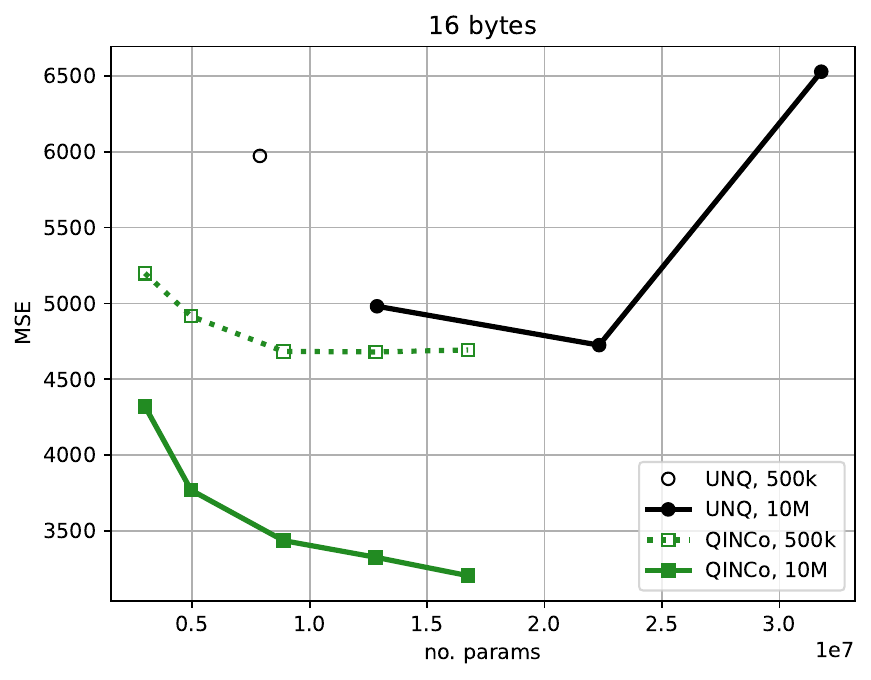}
    \caption{Scaling results comparing UNQ to QINCo. All UNQ models were trained by us using the author's code.
    For the UNQ training with $T\!=\!500$k vectors, all increases in parameter counts based on expanding the encoder/decoder led to overfitting, and so we observed optimal model performance with hyperparameters from the paper. The single point visualized for ``UNQ, 500k'' in both graphs is close to the MSE of the models presented by \citet{Morozov2019UnsupervisedSearch}, but with the model selection criteria outlined in \ref{app:experiments_with_UNQ}.
    For $T\!=\!10$M vectors, we found the best UNQ model used a hidden dimension of $h'\!=\!1,536$ (instead of the default 1,024), and so in our plots we scale the number of layers in the encoder and decoder using $L' \in \{2, 4, 6\}$. Note that these Pareto curves include the optimal performance point for UNQ reported in \cref{tab:scaling_training_vecs_FAISSbaselines}. 
    For \OUR we show curves with $h\!=\!256$ and $L \in \{2,4,8,12,16\}$. 
    With all settings, UNQ  has worse operating points for both model and data scaling than \OUR. In some cases, stability was an issue, as can be seen for the highest parameter count setting with UNQ for the 16-byte results.
     }
    \label{fig:scaling_unq}
\end{figure*}

On 500k training vectors, we found that any increase in model size led to overfitting and increasing MSE numbers. 
However, we did find that UNQ scaled to 10M training vectors quite well for both BigANN1M and Deep1M, with R@1 numbers improving from 34.6\% to 39.7\% and from 26.7\% to 29.2\% on Deep1M, respectively for 8 bytes.
Similar results are observed for 16 bytes.
Despite this, from \cref{fig:scaling_unq} we see that \OUR scales even better; MSE rapidly decreases with increasing capacity with far fewer parameters, for both quantities of training data. This shows that QINCo outperforms UNQ both in the low- and high-data regime (with capacity being scaled accordingly).

Note that we experimented with changing the depth $L'$ of the encoder and decoder of UNQ. This parameter was fixed to $L'=2$ by the authors, and therefore we did not parameterize $L'$ in \cref{tab:complexity}. Including $L'$ in the number of FLOPS for encoding and decoding of UNQ, results in 
$h'\big(D\!+\!(L'-1)h'\!+\!Mb\!+\!MK\big)$ and $h'\big(b\!+\!(L'-1)h'\!+\!D\!+\!M\big)$, respectively.

\begin{table*}
    \centering
    \caption{Performance gain by scaling up from $T\!=\!500$k training vectors to $T\!=\!10$M vectors is limited for OPQ, RQ and LSQ, while \OUR improves further when more training data is available. Also UNQ improves from more training data, see \cref{app:scaling_unq} for more details on scaleability of UNQ. 
    Training on 500k vectors, \OUR is reported with the number of residual blocks $L$ that resulted in best performance.  
    For both rates, this was $L\!=\!12$ for BigANN1M and Deep1M, $L\!=\!1$ for Contriever1M, and $L\!=\!2$ for Fb-ssnpp1M.
    When using  10M training vectors we report \OUR with $L\!=\!16$ in general, and $L\!=\!12$ for Contriever1M.
    For UNQ we report numbers from the original paper~\citep{Morozov2019UnsupervisedSearch}, where models were trained on 500k vectors, as well as the  results of models we trained on 10M vectors using their codebase, denoted UNQ$^*$.
     For the 8-byte setting, UNQ$^*$ achieved highest performance using a hidden dimension of $h'\!=\!1,536$ and $L'\!=\!6$ encoder/decoder layers. For 16 bytes, best performance was found using $h'\!=\!1,536$ and $L'\!=\!4$.
}
       \resizebox{\textwidth}{!}{
    \setlength{\tabcolsep}{3.5pt}
    \begin{tabular}{lccccccccccccccccc}
    \toprule
         \multicolumn{2}{c}{} & 
         \multicolumn{4}{c}{\bf BigANN1M}&  \multicolumn{4}{c}{\bf Deep1M} &\multicolumn{4}{c}{\bf Contriever1M}& \multicolumn{4}{c}{\bf FB-ssnpp1M} \\
         \cmidrule(lr){3-6}\cmidrule(lr){7-10}\cmidrule(lr){11-14}\cmidrule(lr){15-18}
         & $T$ & MSE~($\times 10^4$) & R@1&  R@10& R@100& MSE & R@1&  R@10&  R@100& MSE & R@1&  R@10& R@100&  MSE~($\times 10^4$)& R@1& R@10& R@100 \\
          \midrule
          &\multicolumn{16}{c}{\bf 8 bytes} \\
          \midrule
         OPQ & 500k & 2.95  & 21.9 & 64.8 & 95.4 &  0.26 & 15.9 & 51.2 & 88.2  & 1.87 & 8.0 & 24.7 & 50.8 & 9.52 & 2.5 & 5.1 & 10.9 \\
         OPQ & 10M & 2.99 & 21.3 & 64.3 & 95.6 & 0.26 & 15.1 & 51.1 & 87.9 &  1.87 & 8.5 & 24.3 & 50.4 & 9.52 & 2.5 & 5.0 & 11.2 \\
         RQ & 500k & 2.49 & 27.9 & 75.2 & 98.2 & 0.20 & 21.4 & 63.5 & 95.2 &  1.82 & 10.2 & 26.9 & 52.4 &  9.20 & 2.7 & 6.1 & 13.6 \\
         RQ & 10M & 2.49 & 27.9 & 75.2 & 98.0 & 0.20 & 21.9 & 64.0 & 95.2 & 1.82 & 9.7 & 27.1 & 52.6 &  9.18 & 2.7 & 5.9 & 14.3 \\
         LSQ & 500k & 1.91 & 31.9 & 79.5 & 98.9 & 0.17 & 24.6 & 69.4 & 97.0 &  1.65 & 13.1  & 33.9 & 62.7 & 8.87& 3.3 & 7.5 & 17.3 \\
         LSQ & 10M & 1.89 & 30.6 & 78.7 & 98.9 & 0.17 & 24.5 & 68.8 & 96.7  & 1.64 & 13.1 & 34.9 & 62.5 & 8.82 & 3.5 & 8.0 & 18.2\\
         UNQ & 500k & 1.51 & 34.6 & 82.8 & 99.0 & 0.16 & 26.7 & 72.6 & 97.3 &  --- & --- & --- & --- & --- & --- & --- & ---  \\
        UNQ$^*$ & 10M & 1.12 & 39.7 & 88.3 & 99.6 & 0.14 & 29.2 & 77.5 & 98.8 & --- & --- & --- & --- & --- & --- & --- & --- \\
         \OUR &  500k& 1.38 & 40.2 & 88.0 & 99.6 & 0.15 & 29.4 & 77.6 & 98.5 &  1.57 & 15.4 & 38.0 & 65.5 & 8.95 & 3.0 & 7.7 & 17.1 \\
         \OUR & 10M& \bf 1.12 & \bf 45.2 & \bf 91.2 &\bf  99.7 & \bf 0.12 & \bf 36.3 & \bf 84.6 & \bf 99.4 & \bf 1.40 & \bf 20.7 & \bf 47.4 & \bf 74.6 & \bf8.67 & \bf3.6 & \bf8.9 & \bf20.6 \\
         \midrule
         &\multicolumn{16}{c}{\bf 16 bytes}\\
         \midrule
OPQ & 500k & 1.79 & 40.5 & 89.9 & 99.8  &  0.14 & 34.9 & 82.2 & 98.9 & 1.71 & 18.3 & 40.9 & 65.4 &  7.25 & 5.0 & 11.8 & 25.9 \\
OPQ & 10M  & 1.79 & 41.3 & 89.3 & 99.9  &  0.14 & 34.7 & 81.6 & 98.8 & 1.71 & 18.1 & 40.9 & 65.8 &  7.25 & 5.2 & 12.2 & 27.5 \\
RQ & 500k  & 1.30 & 49.0 & 95.0 & 100.0 &  0.10 & 43.0 & 90.8 & 99.8 & 1.65 & 20.2 & 43.5 & 68.2 &  7.01 & 5.4 & 13.0 & 29.0 \\
RQ & 10M   & 1.30 & 49.1 & 94.9 & 100.0 &  0.10 & 42.7 & 90.5 & 99.9 & 1.65 & 19.7 & 43.8 & 68.6 &  7.00 & 5.1 & 12.9 & 30.2 \\
LSQ & 500k & 0.98 & 51.1 & 95.4 & 100.0 &  0.09 & 42.3 & 89.7 & 99.8 & 1.35 & 25.6 & 53.8 & 78.6 &  6.63 & 6.2 & 14.8 & 32.3 \\ 
LSQ & 10M &  0.97 & 49.8 & 95.3 & 100.0 &  0.09 & 41.4 & 89.3 & 99.8 & 1.33 & 25.8 & 55.0 & 80.1 & \bf6.55 & 6.3 & 16.2 & 35.0 \\
        UNQ & 500k & 0.57 & 59.3 & 98.0 & 100.0 & 0.07 & 47.9 & 93.0 & 99.8 & --- & --- & --- & --- & --- & --- & --- & --- \\
        UNQ$^*$ & 10M & 0.47 & 64.3 & 98.8 & 100.0 & 0.06 & 51.5 & 95.8 & 100.0 & --- & --- & --- & --- &  --- & --- & --- & ---\\
         \OUR &  500k& 0.47 & 65.5 & 99.1 & 100.0 & 0.06 & 53.0 & 96.2 & 100.0 &  1.30 & 26.5 & 54.3 & 79.5 & 6.88 & 5.7 & 14.4 & 31.6 \\
         \OUR  & 10M & \bf 0.32 &\bf  71.9 & \bf 99.6 & \bf 100.0  & \bf 0.05 & \bf 59.8 & \bf 98.0 & \bf 100.0 & {\bf 1.10}  & {\bf 31.1}  & \bf{62.0}& \bf{85.9}   & 6.58 & \bf 6.4 & \bf 16.8 & \bf 35.5 \\ \\
          \bottomrule
    \end{tabular}
    }    \label{tab:scaling_training_vecs_FAISSbaselines}
\end{table*}

\subsection{Dynamic rates}
\label{app:cropping_codes}

\begin{figure*}[ht]
\begin{subfigure}{0.45\linewidth}
\includegraphics[width=\columnwidth]{figs/crop_codes_BigANN_mse.pdf}
\caption{MSE on BigANN1M}
\label{fig:crop_codes_bigann_mse}
\end{subfigure}
\hfill
\begin{subfigure}{0.45\linewidth}
\includegraphics[width=\columnwidth]{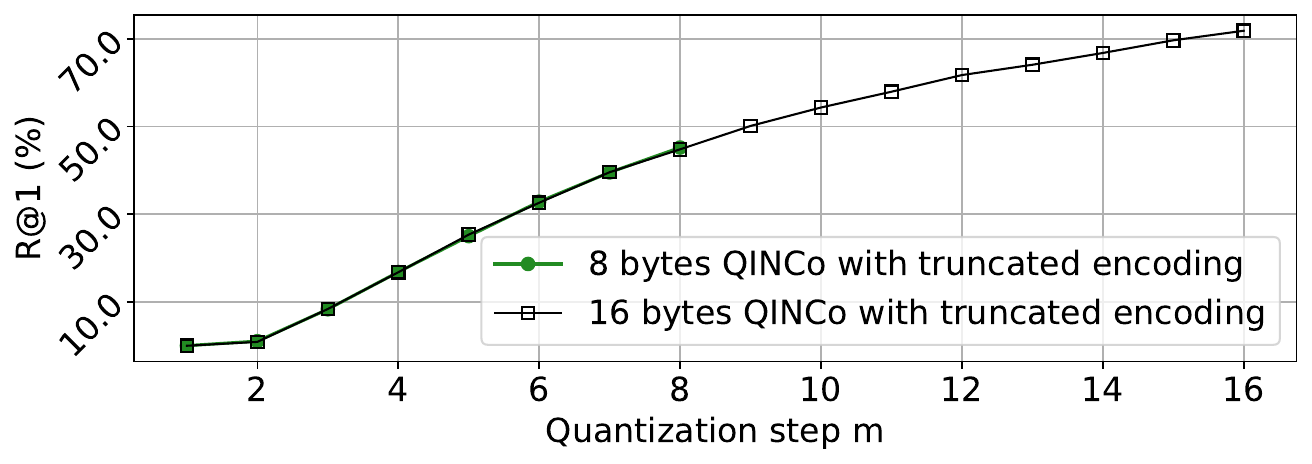}
\caption{R@1 on BigANN1M}
\label{fig:crop_codes_bigann_recall1}
\end{subfigure}
\begin{subfigure}{0.45\linewidth}
\includegraphics[width=\columnwidth]{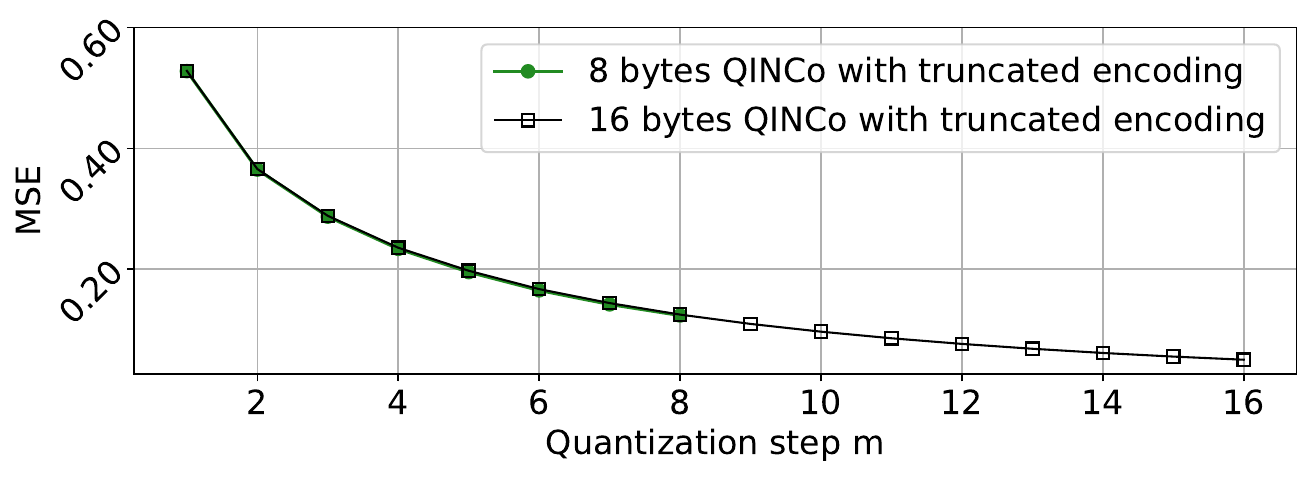}
\caption{MSE on Deep1M}
\label{fig:crop_codes_deep1b_mse}
\end{subfigure}
\hfill
\begin{subfigure}{0.45\linewidth}
\includegraphics[width=\columnwidth]{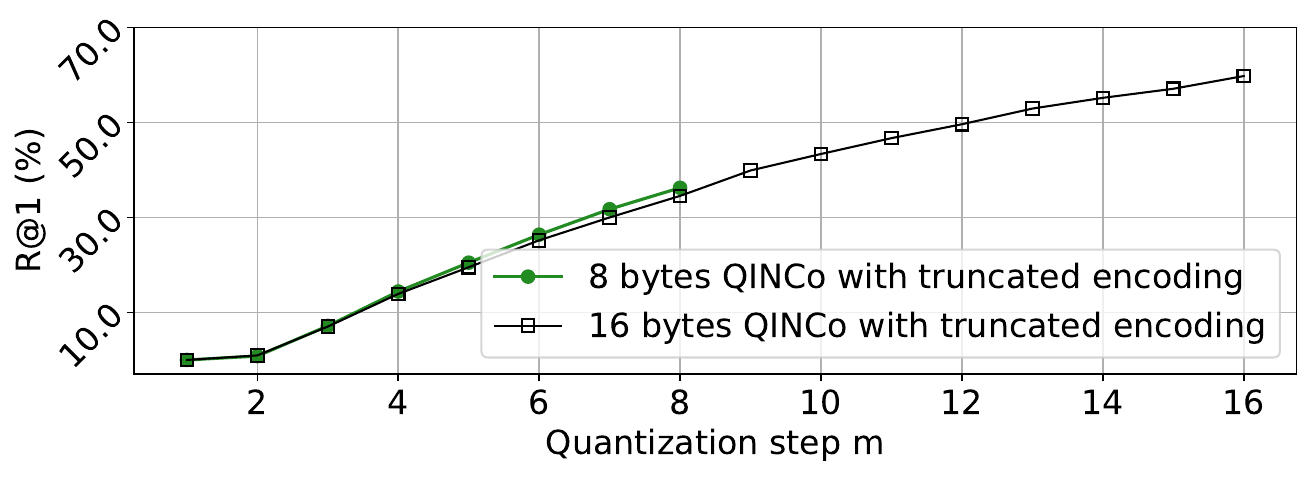}
\caption{R@1 on Deep1M}
\label{fig:crop_codes_deep1b_recall}
\end{subfigure}
\caption{MSE and R@1 for BigANN1M and Deep1M for \OUR ($L=16$) trained for 8-byte and 16-byte encodings, truncated at a varying number of bytes. %
}
\label{fig:cropping_codes_appendix}
\end{figure*}

\Cref{fig:cropping_codes_appendix} shows the MSE and R@1 performance for \OUR trained for 8-byte and 16-byte encoding. 
We observe that \OUR trained for 8- and  16-byte encoding performs very similar at the varying rates.

In \cref{tab:comp_qinco_unq_bitrates} we recap the results of UNQ from \cref{tab:best_db1Mresults} of the main paper using 16-byte encoding, and compare them to \OUR results using 12 and 13 byte encoding.
The results of \OUR using 12 bytes equal or improve over those of UNQ using 16 bytes, except for MSE on Deep1M where \OUR matches UNQ's 16 bytes results with only 13 bytes.

\begin{table}
    \centering
    \caption{Comparison of UNQ with 16-byte encoding, and \OUR with 12- and 13-byte encoding. 
}
    \setlength{\tabcolsep}{3.1pt}
    {\scriptsize
  \begin{tabular}{llcccc}
  \toprule
     &  &
     \multicolumn{2}{c}{\bf BigANN1M}&  \multicolumn{2}{c}{\bf Deep1M}\\
     \cmidrule(lr){3-4}\cmidrule(lr){5-6}
     & Code length & MSE & R@1&  MSE & R@1  \\
     &  & ($\times 10^4$)   \\
     \midrule
UNQ  
     & 16 bytes & 0.57 & 59.3 & 0.07 & 47.9 \\
\OUR & 12 bytes & 0.57 & 61.8 & 0.08 & 49.7 \\
\OUR & 13 bytes & 0.49 & 64.1 & 0.07 & 53.0 \\
    \bottomrule
    \end{tabular}
    }
    \label{tab:comp_qinco_unq_bitrates}
\end{table}

\subsection{Ablations}
\label{app:ablations}

\Cref{tab:ablations} shows results of the ablations for which the main conclusions were provided in \cref{sec:additional_analyzes}. Below we provide more details for each of those.

\mypar{One loss vs $\mathbf{M}$ losses}
\OUR can be trained using only an MSE loss after the last quantization step, \ie $\mathcal{L}^M(\theta)$, instead of using the $M$ losses as given in \cref{eq:loss}. 
In \cref{tab:ablations}, however, we show that this drastically reduces performance. 
Additionally, we observed that optimization became more unstable, which could not be circumvented by using a lower (base) learning rate.

\mypar{Training the $\bf M$ models separately}
The $M$ losses in \OUR can be detached, such that each $m^\text{th}$ loss only updates the trainable parameters in the $m^\text{th}$ part of \OUR. 
\Cref{tab:ablations} shows that MSE in all cases deteriorated, while the recall performances remained rather similar, or slightly increased for 8 bytes Deep1B encoding. In general, we might thus conclude that there is no large effect of the $m^\text{th}$ loss function on earlier quantization steps (i.e. $<m$). 
This corroborates the earlier-made observation that \OUR can be used with dynamic rates during evaluation.

\mypar{Sharing parameters over quantization steps}
The number of trainable parameters in \OUR scales linearly with $M$, the number of bytes used for quantization, see \cref{eq:nr_parameters}. 
To test whether \OUR actually benefits from having $M$ specialized codebook-updating models, we share (a subset of the) parameters of each of those models over all $M$ steps. 
We run three variants: 
(i)   only the parameters of the first concatenation block are shared, 
(ii)  only the parameters of the residual-MLPs are shared, and 
(iii) both the concatenation block and residual-MLP parameters are shared over $M$. 
All models were trained on $T=500$k vectors, and with $L=8$ residual blocks. \Cref{tab:ablations} shows that performance indeed drops when the codebook-predicting models are shared over the $M$ quantization steps. 
A direct relation is visible between the number of parameters that gets reduced by these actions, and the drop in performance. 
This finding suggests that the \OUR benefits from learning $M$ specialized codebook-predicting models.

\begin{table*}[hb]
    \centering
    \caption{Ablation performance for \OUR models trained on $T=500$k vectors, with $L=8$. Compared to the base \OUR model (I), performance heavily degrades when using only the MSE loss on the last quantization step (II). Detaching the $M$ losses does slightly deteriorate the MSE reconstruction performance in all cases, but does not seem to affect recall that much (III). Sharing trainable parameters across the $M$ quantization steps reduces performance (IV-VI), mainly when a large part of the parameters are shared (VI).}
    \resizebox{\textwidth}{!}{
    \setlength{\tabcolsep}{3.5pt}
    \begin{tabular}{llccccccccccc}
    \toprule
         &&  
         \multicolumn{5}{c}{\bf BigANN1M }&  &\multicolumn{5}{c}{\bf Deep1M} \\
         \cmidrule(lr){3-7}\cmidrule(lr){9-13}
         && MSE ($\times 10^4$)&  R@1&  R@10&  R@100& no. params. & & MSE & R@1&  R@10&  R@100 & no. params.\\
          \midrule
          &&\multicolumn{11}{c}{\bf 8 bytes} \\
          \midrule
         I & \OUR & \bf 1.40 & \bf 39.7 & 87.4 &  \bf 99.6 & 4.2M && \bf 0.15 & 29.6 & 77.6 & 98.5 & 3.1M\\
         II & \OUR only last loss $\mathcal{L}^M(\theta)$ & 2.81 & 16.2 & 55.4 & 90.8 & 4.2M && 0.20 & 17.4 & 55.8 & 91.6 & 3.1M  \\ 
         III & \OUR $M$ detached losses & 1.42 & 39.1 & \bf 87.6 & 99.5 & 4.2M && \bf 0.15 & \bf 30.0 & \bf 78.0 & \bf 98.8 & 3.1M\\
         IV & \OUR share concatenate blocks over $M$ & 1.46 & 38.8 & 87.5 & 99.5 & 4.0M && \bf 0.15 & 28.7 & 75.7 & 98.4 & 3.0M\\
         V & \OUR share residual-MLPs over $M$ &1.69 & 37.0 & 85.4 & 99.3 & 1.0M && 0.16 & 27.4 & 74.5 & 98.1 & 0.7M\\
         VI & \OUR share concatenate blocks \& residual-MLPs  &1.66 & 37.1 & 85.2 & 99.4 & 0.8M && 0.16 & 28.4 & 75.4 & 97.9 & 0.6M\\
         \midrule
         &&\multicolumn{11}{c}{\bf 16 bytes}\\
         \midrule
         I & \OUR & \bf 0.47 & 65.7 & \bf 99.0 & \bf 100.0 & 8.9M &&\bf 0.06 & \bf 53.2 &\bf  96.6 & \bf 100.0 & 6.6M \\
        II & \OUR only last loss $\mathcal{L}^M(\theta)$ & 2.85 & 16.1 & 53.2 & 90.1 & 8.9M && 0.14 & 27.1 & 72.3 & 97.1 & 6.6M\\ 
        III & \OUR $M$ detached losses & 0.52 & 65.2 & 98.7 & \bf 100.0 & 8.9M && \bf 0.06 & 53.1 & 96.5 & \bf 100.0 & 6.6M \\
        IV & \OUR share concatenate blocks over $M$ &  0.49 & \bf 66.2 & 99.0 & \bf 100.0 & 8.4M && 0.07 & 51.4 & 95.7 & \bf 100.0 & 6.3M \\
        V & \OUR share residual-MLPs over $M$ & 0.69 & 61.8 & 98.5 & \bf 100.0 & 1.5M && 0.08 & 50.0 & 94.7 & \bf 100.0 & 1.1M\\
        VI & \OUR share concatenate blocks \& residual-MLPs  & 0.71 & 59.4 & 98.3 & \bf 100.0 & 1.1M && 0.08 & 49.6 & 95.2 & \bf 100.0 & 0.8M \\
          \bottomrule
    \end{tabular}
    }
    \label{tab:ablations}
\end{table*}

\end{document}